\documentclass[10pt,twocolumn,letterpaper]{article}

\usepackage{cvpr}
\usepackage{times}
\usepackage{epsfig}
\usepackage{graphicx}
\usepackage{amsmath}
\usepackage{amssymb}
\usepackage{amsfonts}
\usepackage{amssymb}
\usepackage{algorithm}
\usepackage{algorithmic}
\usepackage{multirow}
\usepackage{mathtools}
\usepackage{subfigure}
\usepackage{balance}

\usepackage[pagebackref=true,breaklinks=true,letterpaper=true,colorlinks,bookmarks=false]{hyperref}

\cvprfinalcopy 

\ifcvprfinal\fi
\begin{document}

\title{Maximum Margin Vector Correlation Filter}

\author{Vishnu Naresh Boddeti\\
Robotics Institute\\
Carnegie Mellon University\\
{\tt\small naresh@cmu.edu}
\and
B.V.K.Vijaya Kumar\\
Electrical and Computer Engineering\\
Carnegie Mellon University\\
{\tt\small kumar@ece.cmu.edu}
}

\maketitle

\begin{abstract}
Correlation Filters (CFs) are a class of classifiers which are designed for accurate pattern localization. Traditionally CFs have been used with scalar features only, which limits their ability to be used with vector feature representations like Gabor filter banks, SIFT, HOG \cite{dalal2005histograms}, etc. In this paper we present a new CF named Maximum Margin Vector Correlation Filter (MMVCF) which extends the traditional CF designs to vector features. MMVCF further combines the generalization capability of large margin based classifiers like Support Vector Machines (SVMs) and the localization properties of CFs for better robustness to outliers. We demonstrate the efficacy of MMVCF for object detection and landmark localization on a variety of databases and demonstrate that MMVCF consistently shows improved pattern localization capability in comparison to SVMs.
\end{abstract}

\section{Introduction}
Template-based approaches to image recognition have been popular due to their simplicity and efficiency. These templates are usually designed from multiple training images and they are commonly cross-correlated with the query images to detect patterns in the query. The common approach to template-based methods for recognition tasks is to extract features (e.g., Gabor features \cite{gabor1946theory}, HOG features \cite{dalal2005histograms}, etc.) and build a classifier from these features. Support Vector Machines (SVMs) \cite{boser1992training,cortes1995support} and Correlation Filters (CFs) \cite{kumar2005correlation} are two discriminative template-based classifiers that can be used for pattern detection.

SVMs have been used for many vision tasks such as face detection \cite{osuna1997training}, pedestrian detection \cite{dalal2005histograms} and object detection \cite{felzenszwalb2010object}. Given \textit{N} training vectors $\mathbf{x_i}\in\mathbb{R}^{d}$ and class labels $y_{i}\in\{-1,1\}$ $\forall i\in\{1,\dots,N\}$, the SVM approach finds the hyperplane that maximizes the Euclidean margin (i.e., $l_2$ norm) between the two classes by solving,
\begin{eqnarray}
\min_{\mathbf{f}} &  & \mathbf{f}^{T}\mathbf{f}+C\sum_{i=1}^{N}\xi_{i}\label{eq:svm}\\
 s.t. &  & y_{i}(\mathbf{x_i}^{T}\mathbf{f}+b)\geq1-\xi_{i},\nonumber
\end{eqnarray}
where $\mathbf{f}$ and $b$ represent the hyperplane, $C > 0$ is a trade-off parameter, and $\xi_{i} \geq 0$ is a penalty term. The solution to Eq. \ref{eq:svm} is a linear combination of the training samples, i.e., 
\begin{equation}
\mathbf{f}=\mathbf{X\boldsymbol{\alpha}}
\label{eq:supportvectors}
\end{equation}
where $\mathbf{X}=[\mathbf{x}_{1},\cdots,\mathbf{x}_{N}]$ and the coefficients $\alpha_{i}$ (represented by $\boldsymbol{\alpha}$) being non-zero only for the support vectors.

CFs are a class of classifiers that are most commonly used for pattern detection and are specifically optimized for sliding window based detection. Attractive properties of CFs such as shift-invariance, noise robustness, graceful degradation, and distortion tolerance have been useful in a variety of pattern recognition applications including face detection \cite{bolme2009average}, pedestrian detection \cite{bolme2010simple}, object detection and tracking \cite{bolme2010visual}\cite{rodriguez10automatic_icip}, and biometric classification \cite{SavvidesBiometric}. In this approach a carefully designed template (loosely called a \emph{filter}) $f(p,q)$ is cross-correlated with the query image $t(p,q)$ to produce the output $c(\tau_x,\tau_y)$. This operation can be carried out in the frequency domain taking advantage of the efficiencies afforded by the Fast Fourier Transform (FFT) algorithm, 
\begin{equation}
   \hat{c}=\hat{t}\circ\hat{f}^{*},\label{eq:ifft}
\end{equation}
where $\circ$ is the Hadamard product, $^*$ denotes the complex conjugate operation and $\hat{c}$, $\hat{t}$ and $\hat{f}$ are the 2-D Discrete Fourier transforms (DFTs) of the correlation output, query image and the template, respectively, which can be efficiently implemented via the FFT algorithm. When the query image is from the true-class (i.e., authentic or Class-1), $c(\tau_{x},\tau_{y})$ should exhibit a sharp peak, and when the query image is from a false-class (i.e., impostor or Class-2) $c(\tau_{x},\tau_{y})$ should not have a significant peak. The higher the peak the higher the probability that the query image is from the true-class, and the location of the peak indicates the location of the object. Thus, CFs offer the ability to simultaneously localize and identify objects.

CFs, which have been extensively used for automatic target recognition (ATR) and biometric recognition, have traditionally been used with scalar features (usually raw pixel values or edge maps). However, pixel values (and to an extent edge maps) do not generalize well for object detection in unconstrained environments (e.g., street scenes, indoor scenes, etc.) due to background clutter and substantial variations in color, pose, etc. Discriminative feature representations in conjunction with features that generalize better than pixel values, can provide robustness against these challenges. The performance of the classifier is critically dependent on the choice of the feature representation. Of late, Histogram of Oriented Gradients (HOG) have been shown to perform well on a variety of detection tasks \cite{dalal2005histograms}\cite{felzenszwalb2010object}. When HOG features are extracted from an image, blocks of pixels within the image are transformed to vectors. Thus, the 2-D image is transformed to $K$ 2-D feature channels, where $K$ is the number of feature channels or equivalently the dimension of the vector descriptor at each pixel (or a block of pixels).

Recently Boddeti et.al \cite{boddeti2013correlation} and Kiani et.al \cite{kiani2013multi} proposed a correlation filter design based on ridge regression for vector-valued or multi-channel features which while having very attractive computational and memory efficiencies were also shown to outperform SVMs under the regime of small scale data. However, due to the inherent robustness of SVMs, by way of explicitly maximizing the margin of separation, to outliers they have been shown to outperformed as more and more training data is available. Therefore correlation filters have been shown to outperform SVMs under the regime of small scale data while SVMs begin to outperform correlation filters as more and more data is available. Such an observation has also been made in \cite{rodriguezmaximum} in the context of single channel (scalar/pixels) features. We explicitly address this phenomenon proposing a new classifier design called \textit{Maximum Margin Vector Correlation Filter} (MMVCF) that combines the localization properties of vector-valued correlation filters with the robustness properties of margin maximizing classifiers like SVMs thereby demonstrating localization performance superior to both traditional correlation filters and SVMs under regimes of both small scale and large scale data. The MMVCF design takes into account multiple feature channels while combining the design principles of SVMs and CFs. In contrast to traditional CF designs and SVMs, which treat each feature channel as being independent of each other, the MMVCF design jointly optimizes the performance of multiple channels to produce the desired output by taking advantage of the joint properties of the different feature channels via interactions between the multiple feature channels. The behavior of the CFs of each channel are coordinated across multiple feature channels to exhibit good generalization to unseen patterns by way of a margin maximizing formulation like in SVMs.

MMVCF is equivalent to an SVM in a transformed space (shown in Section 3), or equivalently MMVCF maximizes a non-Euclidean margin. Sivaswamy et al. \cite{shivaswamy21relative} recently showed that the \emph{type }of margin that should be maximized is important while designing maximum margin classifiers. For example, Ashraf et al. \cite{ashraf2010re} maximized a non-Euclidean margin for their task to apply Gabor filters in a lower dimensional feature space. The proposed classifier maximizes a non-Euclidean margin, but our solution is motivated by criteria for precise object localization.

Many methods for learning filters in a convolutional framework have been proposed like \cite{rigamonti2011filter} and convolutional neural networks \cite{lecun1995convolutional} and many recent convolution based sparse coding methods \cite{yang2010supervised}. While all such methods learn filters in a convolutional framework, the filters are optimized for minimizing reconstruction error of image patches instead of pattern localization. Such methods learn those filters for feature representation rather than convolutional pattern detection like correlation filters.

Thornton et al. \cite{thornton2005linear} proposed what they called SVM Correlation Filter, but their work is very different from ours. Firstly, it was designed for scalar feature representations and secondly, they adopt a brute force approach by simply treating shifted versions of the true-class images as the virtual false-class samples, which does not scale well with the number of training images and the dimensionality of the image.

\section{Background}
 CF is a spatial-frequency array (equivalently, a template in the image domain) that is specifically designed from a set of training patterns that are representative of a particular pattern class. CFs primarily seek to explicitly control the shape of the entire cross-correlation output between the image and the filter unlike other classifiers (e.g., SVMs) which only control the output value at the target location. Towards this end many CF designs \cite{mahalanobis1994unconstrained}\cite{bolme2009average}\cite{bolme2010visual}, all of them assuming scalar features at every pixel, have been proposed which minimize the Mean Square Error (MSE) between the ideal desired correlation output for a true-class (or false-class) input image and the cross-correlation output of the training images with the filter. Given $N$ training images, the filter design problem is posed as an optimization problem (for notational ease, expressions are given for 1-D signals),
\begin{eqnarray}
\label{eq:mosse}
\min_{\mathbf{f}} && \frac{1}{N}\sum_{i=1}^{N}\|\mathbf{x_i}\otimes\mathbf{f} - \mathbf{g_i}\|^2_2 + \lambda \|\mathbf{f}\|^2_2
\end{eqnarray}
where $\otimes$ denotes the cross-correlation operation, $\mathbf{x_i}$ denotes the $i-$th image, $\mathbf{f}$ denotes the CF template and $\mathbf{g_i}$ denotes the desired correlation output for the $i-$th image, and $\lambda$ is the regularization parameter. To achieve good object localization, the CFs are usually designed to give a sharp peak at the center of the correlation output plane for a centered true-class pattern and no such peak for a false-class pattern (for example, $\mathbf{g_i} = [0,\dots,0,1,0,\dots,0]^T$ for true class and $\mathbf{g_i} = [0,\dots,0,-0.1,0,\dots,0]^T$ for false class). In addition to minimizing the localization loss, some filter designs \cite{mahalanobis1987mace}\cite{kumar1994optimal} also constrain the output at the target location,
\begin{eqnarray}
\label{eq:otsdf}
\min_{\mathbf{f}} && \frac{1}{N}\sum_{i=1}^{N}\|\mathbf{x_i}\otimes\mathbf{f} - \mathbf{g_i}\|^2_2 + \lambda \|\mathbf{f}\|^2_2 \\
s.t. && \mathbf{f}^T\mathbf{x_i} = q_i \nonumber
\end{eqnarray}
where $q_i$ is the desired output value at the target location.

\section{Maximum Margin Vector Correlation Filter}
Traditional CFs have been often designed using scalar features (most commonly pixel values) and hence cannot be directly used with vector features like HOG features which are represented as $K$-dimensional vector functions, where $K$ denotes the number of feature channels ($K=32$ in this paper for HOG features as in \cite{felzenszwalb2010object}). Recently an unconstrained ridge regression based method has been proposed for designing correlation filters with vector-valued or multi-channel features, henceforth referred to as Vector Correlation Filters (VCF), which while outperforming SVMs for localization tasks like car and face alignment, pedestrian and car detection can suffer from poor robustness to outliers. On the other hand SVMs due to their margin maximizing property are more robust to outliers and noisy data. Therefore by combining the localization loss of the correlation filter with the hinge loss of the SVMs we can improve the localization capability of SVMs and the generalization capability of VCF. We refer to the resulting classifier design as Maximum Margin Vector Correlation Filter (MMVCF).
\begin{figure}[!h]
\begin{centering}
\includegraphics[scale=0.055]{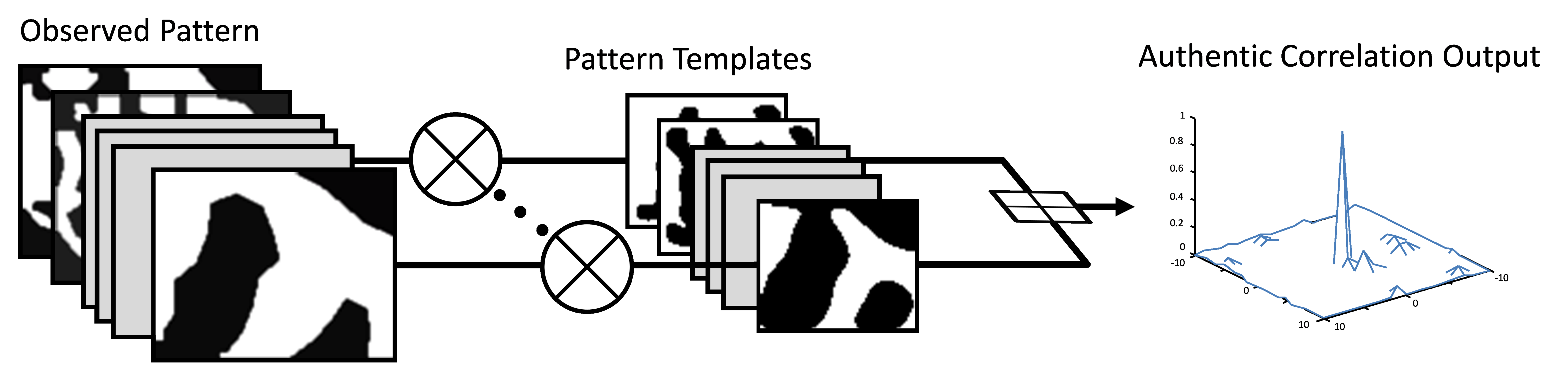}
\par\end{centering}
\caption{The outputs of each feature channel are aggregated to compute the final correlation output which is desired to have a sharp peak at the object location for the correct object class.}
\label{fig:vector-OTSDF-correlation}
\end{figure}
MMVCF consists of one CF per feature channel which are optimized to minimize the localization loss defined as the MSE between the correlation output and the desired ideal correlation output. Since each feature (corresponding to each branch, see Fig. \ref{fig:vector-OTSDF-correlation} for a pictorial description of MMVCF) leads to a peak (at least for the correct object) at the same location, the final output can be obtained by coherently adding all the branch outputs. However, due to the summation, the final correlation output plane is not necessarily optimized as it is in the case of an individual CF like in Eq. \ref{eq:mosse} or Eq. \ref{eq:otsdf}. As opposed to individual CF design, the vector feature design takes advantage of the joint properties of different feature channels which results in the optimal correlation output plane. Hence the MMVCF detector allows for more degrees of freedom to satisfy the CF design criteria leading to robust discrimination capabilities. We design all $K$ CFs jointly, such that the sum of their outputs satisfies our design criteria. The MMVCF design for $N$ training images is formulated as the following optimization problem,

{\tiny
\begin{eqnarray}
  \min_{\mathbf{f^1},\mathbf{f^2},\dots,\mathbf{f^K}} && \frac{1}{N}\sum_{i=1}^N\left\|\sum_{k=1}^K\mathbf{x^k_i}\otimes\mathbf{f^k}-\mathbf{g_i}\right\|^2_2 + \lambda\sum_{k=1}^K\left\|\mathbf{f^k}\right\|^2_2 + C\sum_{i=1}^N\xi_i \\
  s.t. && y_i\sum_{k=1}^K\mathbf{f^{kT}}\mathbf{x^k_i} \geq q_i - \xi_i \nonumber
\end{eqnarray}}
where $q_i$ is the response for sample $\mathbf{x}_i$, $\lambda \geq 0$ is the regularization parameter,  $C > 0$ is a trade-off parameters, $\xi_i$ is a penalty term, the feature $\mathbf{x}$ and filter $\mathbf{f}$ are represented by their $K$-channels i.e., $\mathbf{x} = \{\mathbf{x}^1,\mathbf{x}^2,\dots,\mathbf{x}^K\}$ and $\mathbf{f} = \{\mathbf{f}^1,\mathbf{f}^2,\dots,\mathbf{f}^K\}$. Using Parseval's Theorem \cite{oppenheim1997signals} the above optimization problem can be posed equivalently in the frequency domain resulting in a closed form expression for the objective. Further since inner products are preserved between the spatial and frequency domains the constraints can also be mapped into the frequency domain. This results in an efficient solution for the MMVCF,

{\tiny
\begin{eqnarray}
\label{eq:vcf_freq}
  \min_{\mathbf{\hat{f}^1},\mathbf{\hat{f}^2},\dots,\mathbf{\hat{f}^K}} && \frac{1}{N}\sum_{i=1}^N\left\|\sum_{k=1}^K\mathbf{\hat{X}^{k\dagger}_i}\mathbf{\hat{f}^k}-\mathbf{\hat{g}_i}\right\|^2_2 + \lambda\sum_{k=1}^K\left\|\mathbf{\hat{f}^k}\right\|^2_2 + C\sum_{i=1}^N\xi_i \\
  s.t. && y_i\sum_{k=1}^K\mathbf{\hat{f}^{k\dagger}}\mathbf{\hat{x}^k_i} \geq q_i - \xi_i \nonumber
\end{eqnarray}}
$\mathbf{\hat{x}}$ denotes the Fourier transform of $\mathbf{x}$ and $\mathbf{\hat{X}}$ denotes a diagonal matrix whose diagonal entries are the elements of $\mathbf{\hat{x}}$ and $\dagger$ denotes the conjugate transpose operation. We compute the frequency domain representation of $\mathbf{x}$ by computing the Fourier transform of its $K$ channels independently i.e., $\mathbf{\hat{x}} = \{\mathbf{\hat{x}}^1,\mathbf{\hat{x}}^2,\dots,\mathbf{\hat{x}}^K\}$. Further, we set the desired ideal CF to a scaled Gaussian to match with the inequality constraints at the target location i.e., $\mathbf{g_i} = \mathbf{\tilde{g}}\times\sum_{k=1}^K\mathbf{f}^{kT}\mathbf{x^k_i}$, where $\tilde{g}(n) = \exp\left(-\frac{(n-\mu)^2}{2\sigma^2}\right)$ with $\mu$ being the object location. For a $d$ dimensional input, the objective function in Eq. \ref{eq:vcf_freq} can be reduced to the following quadratic function,
\begin{eqnarray}
\label{eq:quadratic}
  \min_{\mathbf{\hat{f}}} && \mathbf{\hat{f}^{\dagger}}\mathbf{\hat{S}}\mathbf{\hat{f}} + C\sum_{i=1}^N \xi_i\\
  s.t. && y_i\mathbf{\hat{f}}^{\dagger}\mathbf{\hat{x}_i} \geq q_i - \xi_i \nonumber
\end{eqnarray}
\noindent where $\mathbf{\hat{S}} = \mathbf{\hat{D}} + \lambda\mathbf{I} - \mathbf{\hat{P}}$, with $\mathbf{I}$ being an identity matrix of appropriate dimensions, and 
\begin{equation}
\mathbf{\hat{D}}=\left[\begin{array}{ccc}
\frac{1}{N}\sum_{i=1}^N\mathbf{\hat{X}_{i}^{1\dagger}}\mathbf{\hat{X}_{i}^{1}} & \cdots & \frac{1}{N}\sum_{i=1}^N\mathbf{\hat{X}_{i}^{1\dagger}}\mathbf{\hat{X}_{i}^{k}}\\
\vdots & \ddots & \vdots\\
\frac{1}{N}\sum_{i=1}^N\mathbf{\hat{X}_{i}^{k\dagger}}\mathbf{\hat{X}_{i}^{1}} & \cdots & \frac{1}{N}\sum_{i=1}^N\mathbf{\hat{X}_{i}^{k\dagger}}\mathbf{\hat{X}_{i}^{k}}
\end{array}\right]
\end{equation}
\begin{equation}
\mathbf{\hat{P}}=\left[\begin{array}{c}
\frac{1}{dN}\sum_{i=1}^N\mathbf{\hat{X}^1_i\hat{g}_i\hat{x}^{1\dagger}_i}\\
\vdots\\
\frac{1}{dN}\sum_{i=1}^N\mathbf{\hat{X}^k_i\hat{g}_i\hat{x}^{k\dagger}_i}\\
\end{array}\right]\: \mathbf{\hat{f}}=\left[\begin{array}{c}
\mathbf{\hat{f}^1}\\
\vdots\\
\mathbf{\hat{f}^{k}}
\end{array}\right]\:
 \mathbf{\hat{x}_i}=\left[\begin{array}{c}
\mathbf{\hat{x}^1_i}\\
\vdots\\
\mathbf{\hat{x}^{k}_i}
\end{array}\right] \nonumber
\end{equation}
where $\mathbf{\hat{D}}$ is the cross-power spectrum matrix (interaction energy between the feature channels). The parameter $\lambda$ offers a trade-off between the localization loss and the $l_2$ regularization. In order to use a bounded parameter (for implementation purposes), we weight both terms as, $\mathbf{\hat{S}} = (1-\gamma)(\mathbf{\hat{D}} - \mathbf{\hat{P}})+ \gamma\mathbf{I}$, where $0\leq \gamma \leq 1$. Setting $\gamma=1$ will ignore the localization criterion and result in the regular SVM classifier for registered images and smaller values of $\gamma$ can improve object localization by forcing sharper peaks in the correlation plane. Since $\hat{\mathbf{S}}$ is a positive definite matrix, we can transform the data such that $\mathbf{\tilde{x}}_{i}=\hat{\mathbf{S}}^{-\frac{1}{2}}\hat{\mathbf{x}}_{i}$ and $\mathbf{\tilde{f}}=\hat{\mathbf{S}}^{\frac{1}{2}}\hat{\mathbf{f}}$ and rewrite the criterion as,
\begin{eqnarray}
\label{eq:prewhitened}
\min_{\mathbf{\tilde{f}}} &  & \mathbf{\tilde{f}^{\dagger}\tilde{f}}+C\sum_{i=1}^{N}\xi_{i}\\
s.t. &  & y_i\mathbf{\tilde{f}}^{\dagger}\mathbf{\tilde{x}_i} \geq q_i-\xi_{i}.\nonumber
\end{eqnarray}
\noindent The dual formulation of the problem in Eq. \ref{eq:quadratic} is,
{\small
\begin{eqnarray}
\label{eq:mmvcfdual}
\max_{0\leq\alpha_{i}\leq C} && \sum_{i=1}^{N}q_i\alpha_{i}-\frac{1}{2}\sum_{i=1}^{N}\sum_{j=1}^{N}\alpha_{i}\alpha_{j}y_{i}y_{j}K(\mathbf{\hat{x}}_{i},\mathbf{\hat{x}}_{j})\\
s.t. && \sum_{i=1}^{N}y_{i}\alpha_{i} = 0 \nonumber 
\end{eqnarray}}
where $K(\mathbf{\hat{x}}_{i},\mathbf{\hat{x}}_{j})=\mathbf{\hat{x}^{\dagger}_{i}}\mathbf{\hat{S}^{-1}\hat{x}_{j}}$ is the kernel matrix (defined in the frequency domain) which gives geometric shift invariance to the classifier (up to the cell size in the context of HOG feature representation).

\section{Implementation Issues}
The MMVCF design can be implemented using a standard SVM solver by solving either the primal formulation in Eq. \ref{eq:prewhitened} using the transformed images to find $\mathbf{\tilde{w}}$ or by solving the dual formulation in Eq. \ref{eq:mmvcfdual} to compute the coefficients $\boldsymbol{\alpha}$. Solving the dual problem in Eq. \ref{eq:mmvcfdual} requires us to compute $\mathbf{\hat{x}_i^{\dagger}}\mathbf{\hat{S}^{-1}\hat{x}_j}$. Since $\mathbf{\hat{S}}$ is a non-diagonal matrix, naively inverting it is computationally expensive. The ``localization loss" term $\mathbf{\hat{S}}$ which is composed of $\mathbf{\hat{D}}$ and $\mathbf{\hat{P}}$ can be approximated by ignoring $\mathbf{\hat{P}}$ since the entries in $\mathbf{\hat{S}}$ are smaller than those in $\mathbf{\hat{D}}$ by a factor of $d$. Ignoring $\mathbf{\hat{P}}$ is equivalent to minimizing the energy of the entire correlation plane, including the correlation value at the target location. The contribution of the correlation value at the target to the energy of the correlation plane is negligible, and therefore does not adversely affect the filter solution. This approximation allows us to take advantage of the unique structure of $\mathbf{\hat{D}}$, i.e., a sparse block matrix structure where each block is a diagonal matrix, for efficiently computing its inverse by a block-wise matrix inversion. Empirically it was observed that using this approximation results in a negligible loss in filter performance in our experiments. 

During test time the $K$-channel representation of the filter $\mathbf{f} = \{\mathbf{f}^1,\mathbf{f}^2,\dots,\mathbf{f}^K\}$ is applied to a $K$-channel representation of an image $\mathbf{x} = \{\mathbf{x}^1,\mathbf{x}^2,\dots,\mathbf{x}^K\}$ by cross correlating each feature channel filter $\mathbf{f}^k$ with its corresponding feature channel $\mathbf{x}^k$ and finally summing up all the feature channel outputs. For efficiency, the cross-correlations are performed in the frequency domain via FFTs.

\section{Experiments}
To demonstrate the efficacy of the proposed classifier, we evaluate its performance over a number of different databases for object detection and object part localization under the regimes of both small scale and large scale data. For each of these databases we compare the performance of the proposed MMVCF, VCF and SVM. For all the datasets, the images are represented using the HOG features as implemented in \cite{PMT}, and object detection is done by cross-correlating the template (i.e., represented by the $\mathbf{f}$ described above) with the HOG feature representation of the query image at multiple scales via a pyramid approach following \cite{felzenszwalb2010object}. In addition, we applied the retraining technique described by Dalal and Triggs \cite{dalal2005histograms}, i.e., we iteratively apply the filter to the \emph{training} frames and add the false positives as false-class images. The computation required to test any of these filters on a given image is exactly the same, so no computational comparison is given. Further the best parameters for SVM, VCF and MMVCF are estimated by cross-validation on separate validation sets. Typically small values of $\gamma = \{10^{-2}, 10^{-1}\}$ are best for MMVCF while $\gamma=1$ corresponds to an SVM. 

\subsection{Pedestrian Detection}
\begin{figure}[!h]
	\centering
	\includegraphics[scale=0.4]{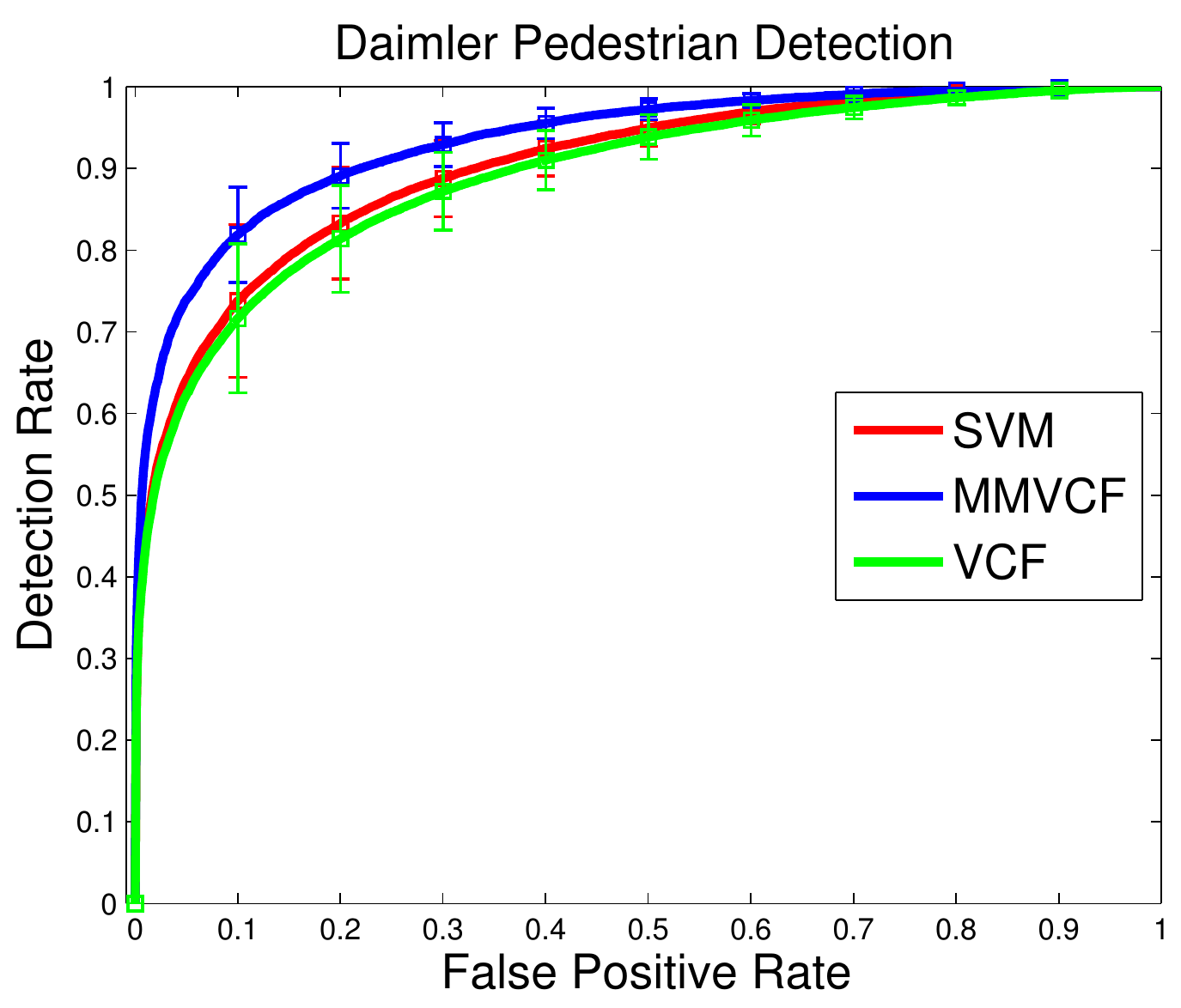}
	\caption{ROC curve of detection rate vs false positive rate comparing MMVCF, SVM and VCF for pedestrian detection}
	\label{fig:daimler}
\end{figure}

We evaluated our method for pedestrian detection using Daimler pedestrian dataset \cite{munder2006experimental} containing five disjoint images sets, three for training and two for testing. Each set consists of 4800 pedestrian and 5000 non-pedestrian images of size $36\times18$. We compute HOG features using 5 orientation bins with cell and block sizes of $3\times3$. We train MMVCF, VCF and SVM using all the negative and positive training samples. Given a test image, we first correlate it with the trained detectors and then measure the peak sharpness via the Peak-to-Sidelobe Ratio, ratio of peak response to response of surrounding region (see \cite{kumar2005correlation} for details). We follow the protocol described in \cite{munder2006experimental} to report our results and to cross-validate over the parameters for VCF, SVM and MMVCF. Figure.\ref{fig:daimler} shows the full ROC curves for pedestrian detection while Table.\ref{table:daimler} shows the mean and the standard deviation of the area under the curve for each of the three detectors that we are comparing. This is a medium sized dataset where the performance of VCF and SVM are nearly the same (VCF outperforms SVM when using fewer training samples \cite{kiani2013multi} on this dataset) while MMVCF outperforms both SVM and VCF.
\begin{table}[!h]
      \centering
      \caption{Daimler Pedestrian: AUC (mean and deviation)}
      \label{table:daimler}
      \begin{tabular}{|c|c|c|c|c|}
      \hline
      AUC & {\footnotesize VCF} & {\footnotesize SVM($\gamma=1$)} & {\footnotesize MMVCF}\\
      \hline
      (in \%) & {\footnotesize 89.1 (4.3)} & {\footnotesize 90.2 (3.9)} & {\footnotesize 93.4 (2.3)} \\
      \hline
      \end{tabular}
\end{table}
\subsection{Object Alignment}
Since MMVCFs are designed for accurate localization of object parts we consider the task of multi-view car alignment from a single image \cite{li2009robust}\cite{boddeti2013correlation}. This is a challenging task since most car parts are only weakly discriminative for detection and the appearance of the cars can change dramatically as the viewing angle changes. Further cars in natural street scenes vary widely in shape and are often present in highly cluttered backgrounds, with severe occlusion, self or otherwise, in many instances. VCFs have been shown to perform well on this task and we compare VCF, SVM and MMVCF based landmark detectors for the appearance model while using the robust shape model introduced in \cite{li2009robust} by Li et.al. due to its ability to handle gross landmark detection errors caused either by partial occlusions or clutter in the background.

We evaluate the proposed approach on cars from the MIT Street Dataset \cite{mit-street-dataset} which contains over 3500 street scene images created for the task of object recognition and scene understanding. This dataset has annotated landmarks for 3,433 cars spanning a wide variety of types, sizes, backgrounds and lighting conditions including partial occlusions. All the shapes are normalized to roughly a size of $250\times130$ by Generalized Procrustes Analysis \cite{procrustes}. The dataset is manually classified into five different views and due to space constraints we compare the landmark detectors on 1400 images of the half-frontal view since this view has the most amount of shape variation and number (14) of visible points. We randomly selected 400 images from each view for training and use the rest of the images for testing. Patches from occluded landmarks are excluded while training the part detectors and for evaluation the occluded landmark is placed at the most likely location in the image.

For each landmark, we extract a $96\times96$ image patch as the positive sample and negative samples of the same size are extracted uniformly around each landmark. Each of these local patches are further represented by the Histogram of Oriented Gradients (HOG) descriptor. The HOG descriptors are computed over dense and overlapping grids of spatial blocks, with image gradient features extracted at 9 orientations and a spatial bin size of $4\times4$. The Linear SVM, VCF and the proposed MMVCF are designed using these HOG representations of the patches. 

\begin{figure}[!ht]
	\centering
	\subfigure[]{\includegraphics[scale=0.22]{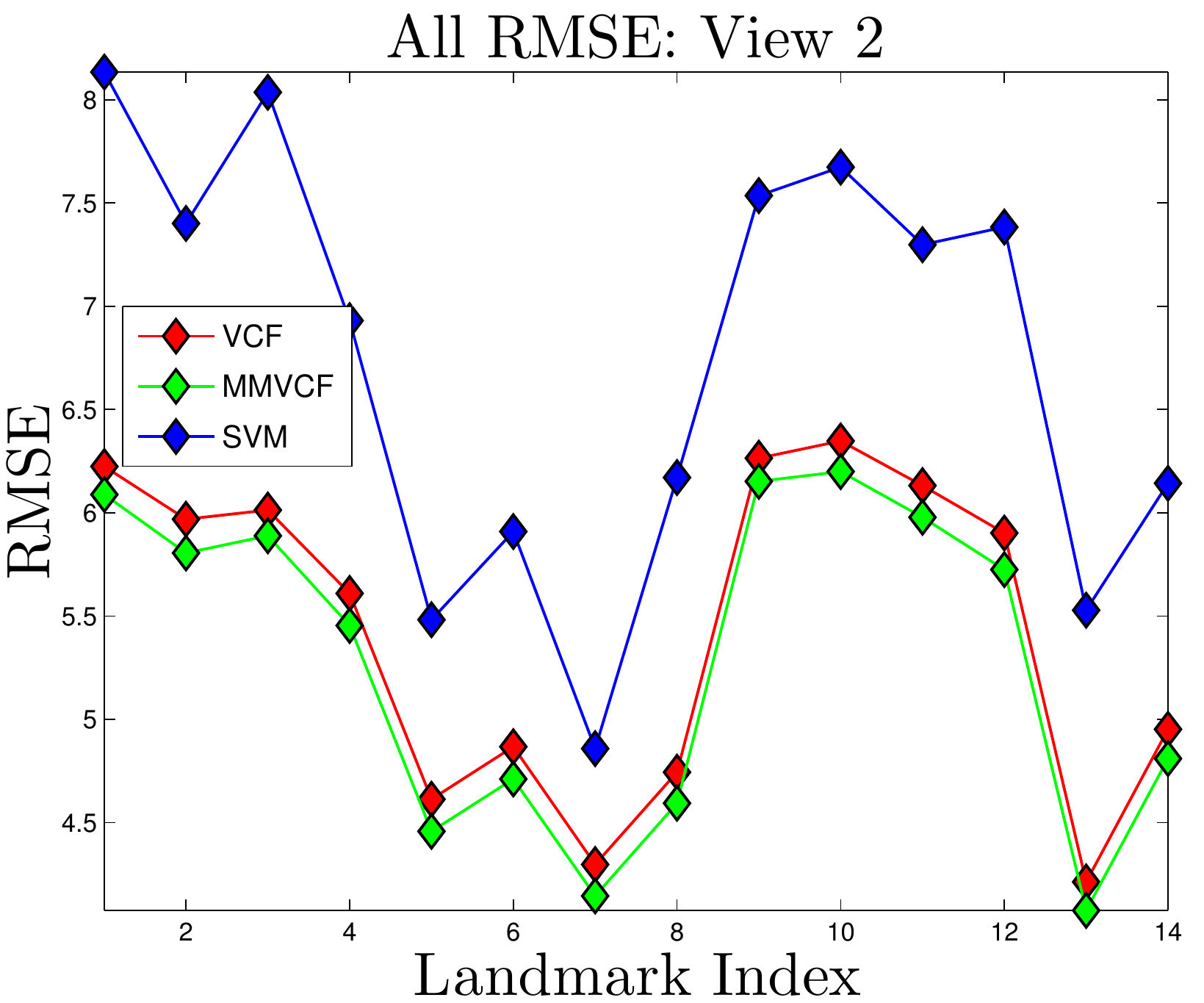}}
	\subfigure[]{\includegraphics[scale=0.22]{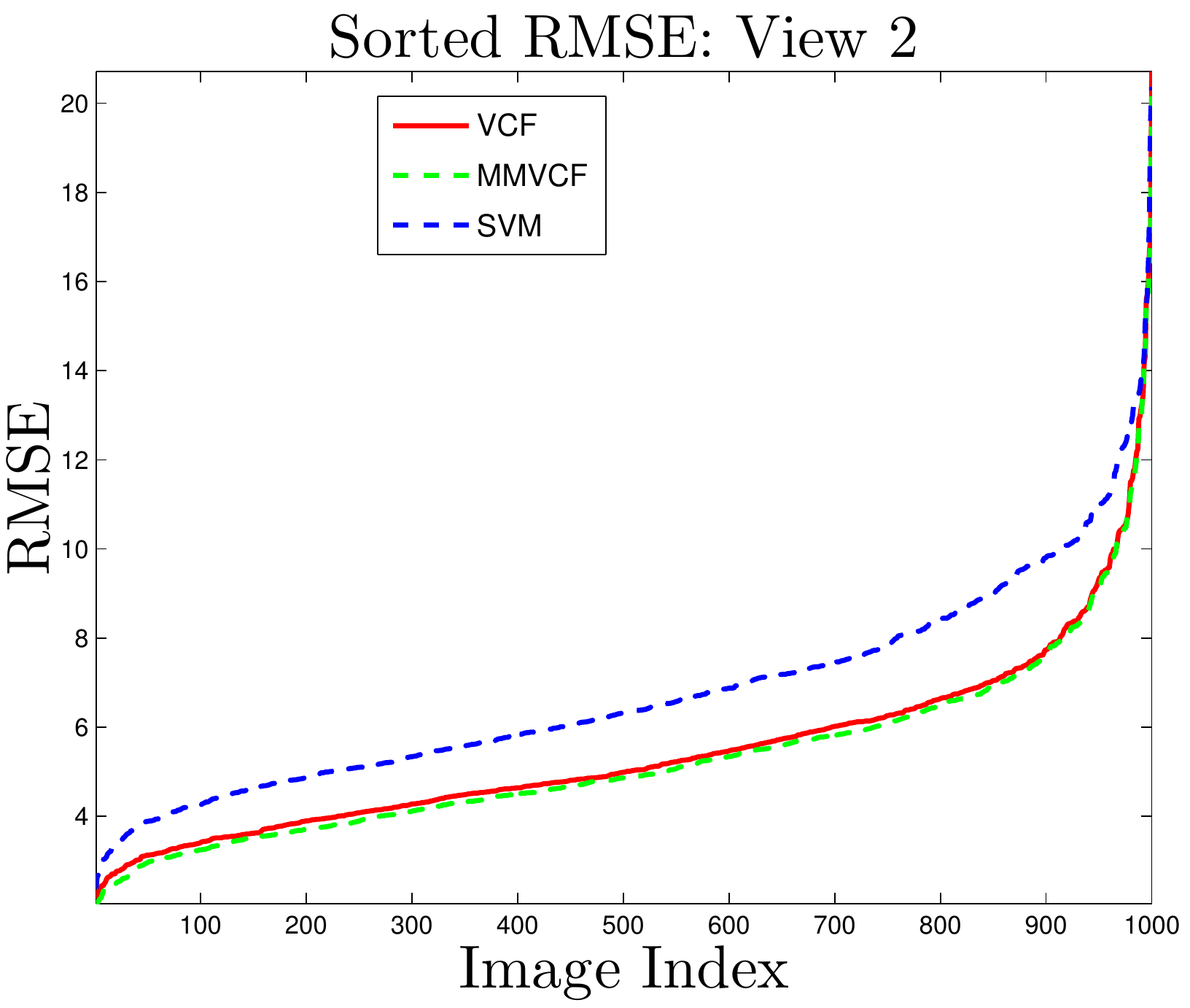}}
	\caption{a) RMSE of 14 landmarks averaged over all images. b) Comparison of sorted RMSE for different landmark detectors.}
	\label{fig:alignment_error}
\end{figure}

\begin{figure}[!ht]
	\includegraphics[trim = 11cm 10cm 12cm 10cm,clip=true,totalheight=0.072\textheight]{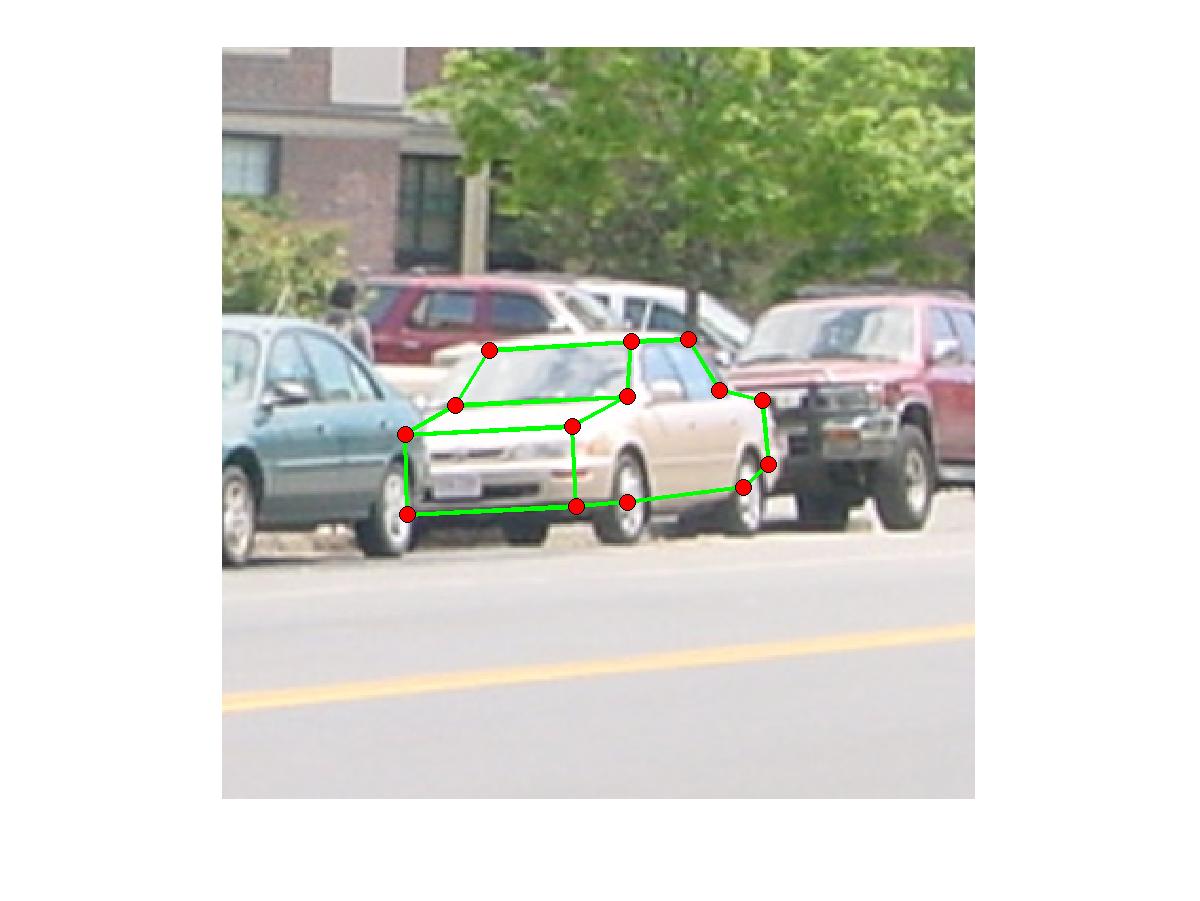}	
	\includegraphics[trim = 11cm 10cm 12cm 10cm,clip=true,totalheight=0.072\textheight]{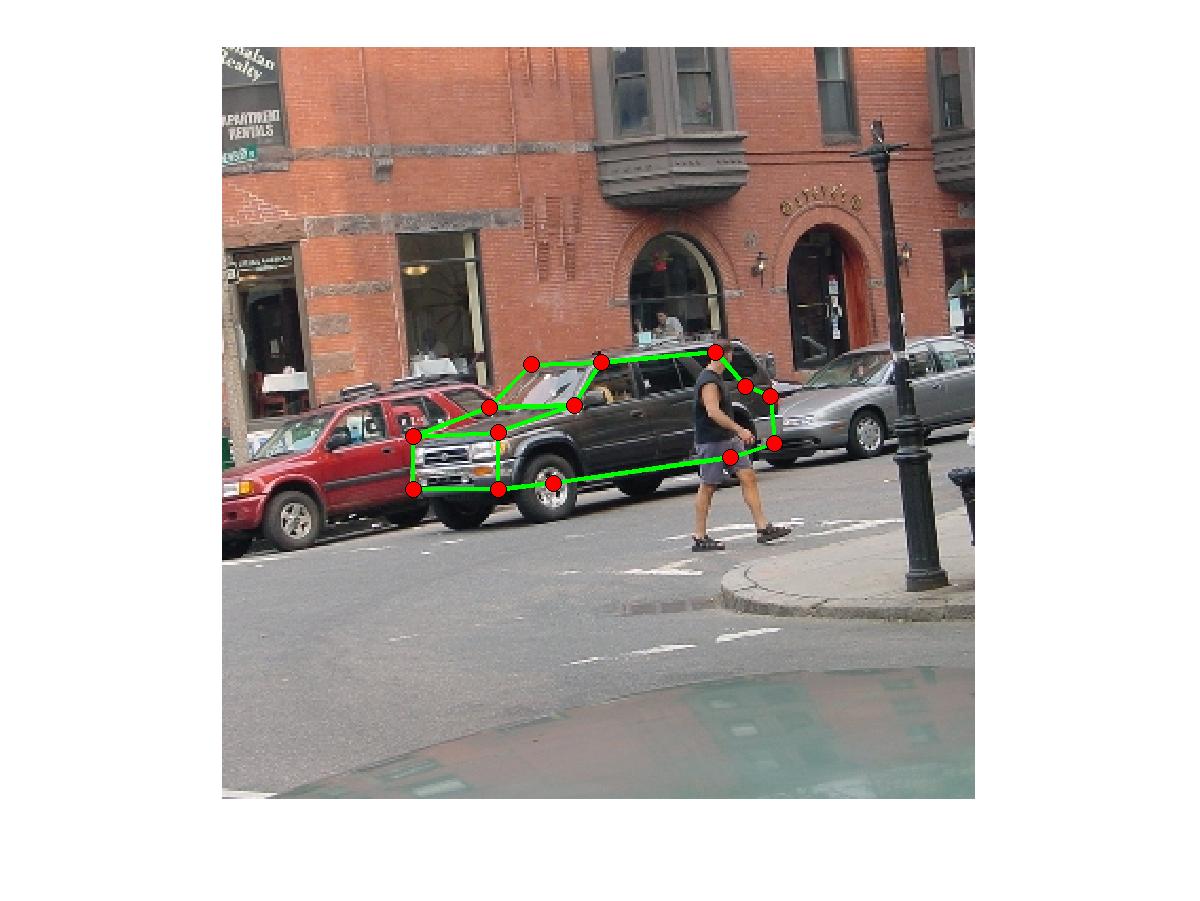}
	\includegraphics[trim = 11cm 10cm 12cm 10cm,clip=true,totalheight=0.072\textheight]{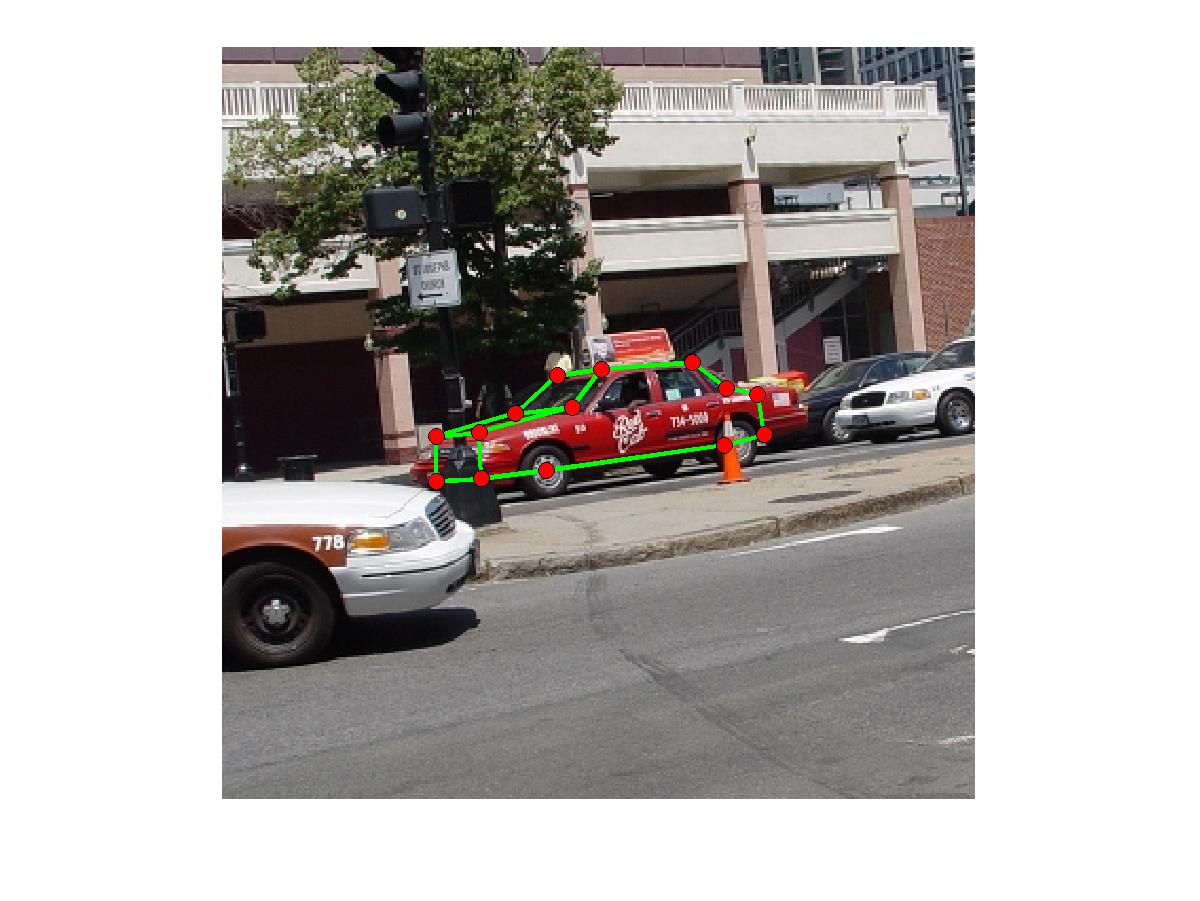}\\
	\includegraphics[trim = 11cm 10cm 12cm 10cm,clip=true,totalheight=0.072\textheight]{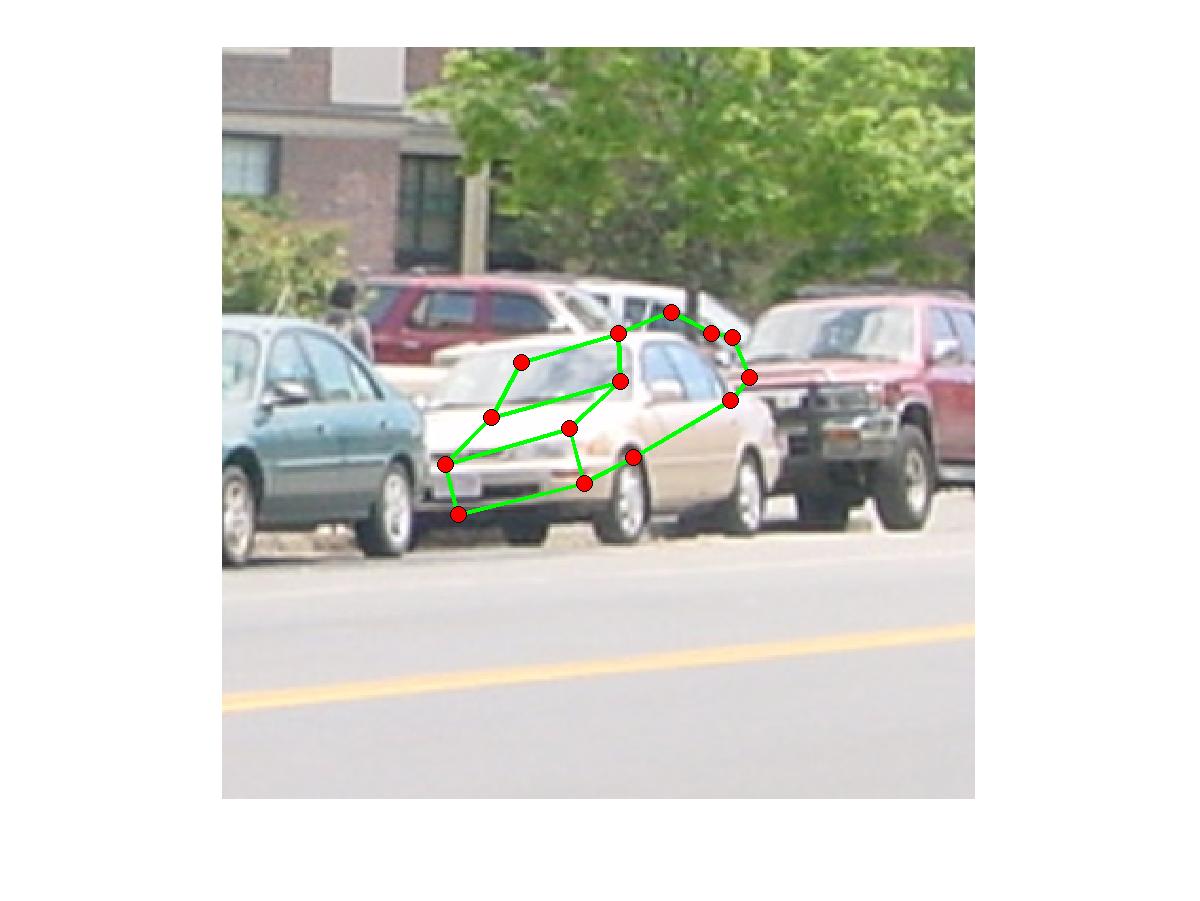}
	\includegraphics[trim = 11cm 10cm 12cm 10cm,clip=true,totalheight=0.072\textheight]{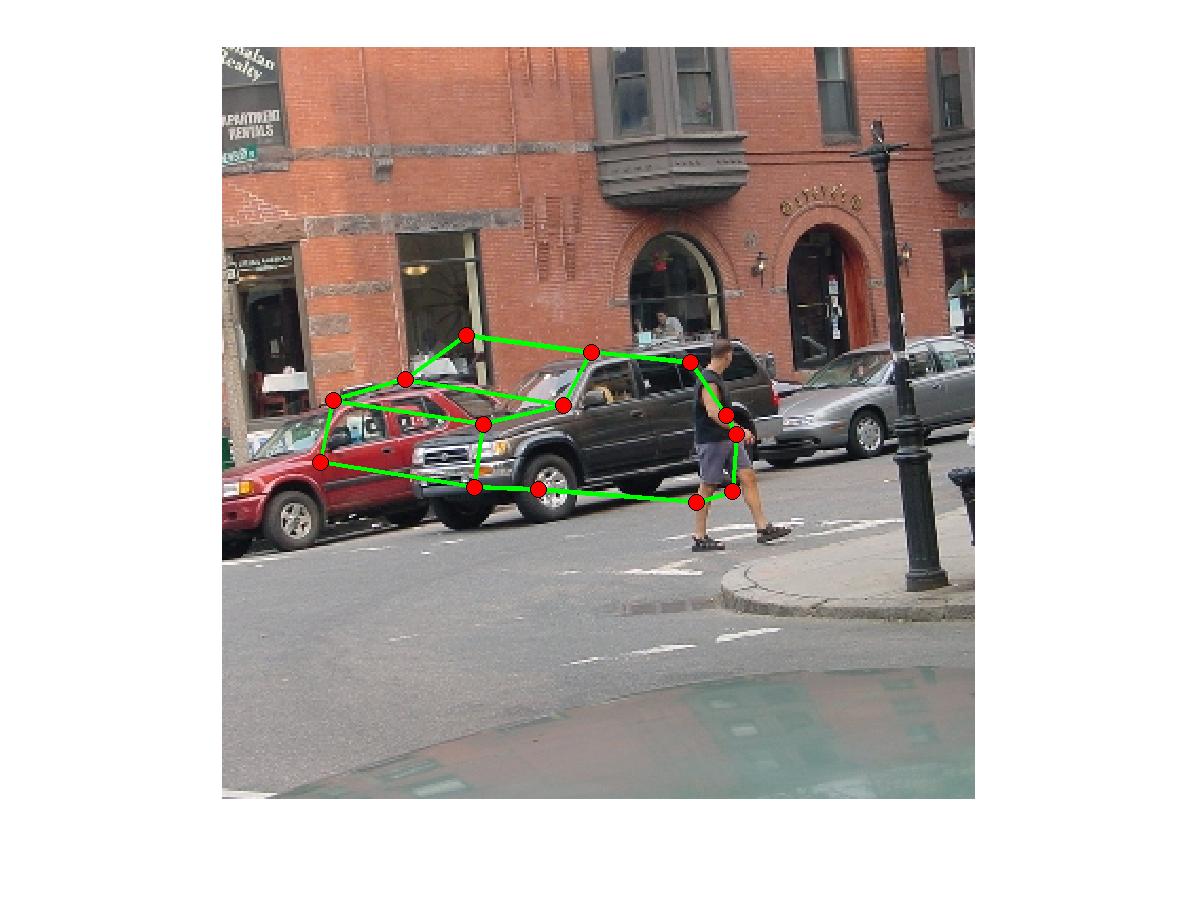}
	\includegraphics[trim = 11cm 10cm 12cm 10cm,clip=true,totalheight=0.072\textheight]{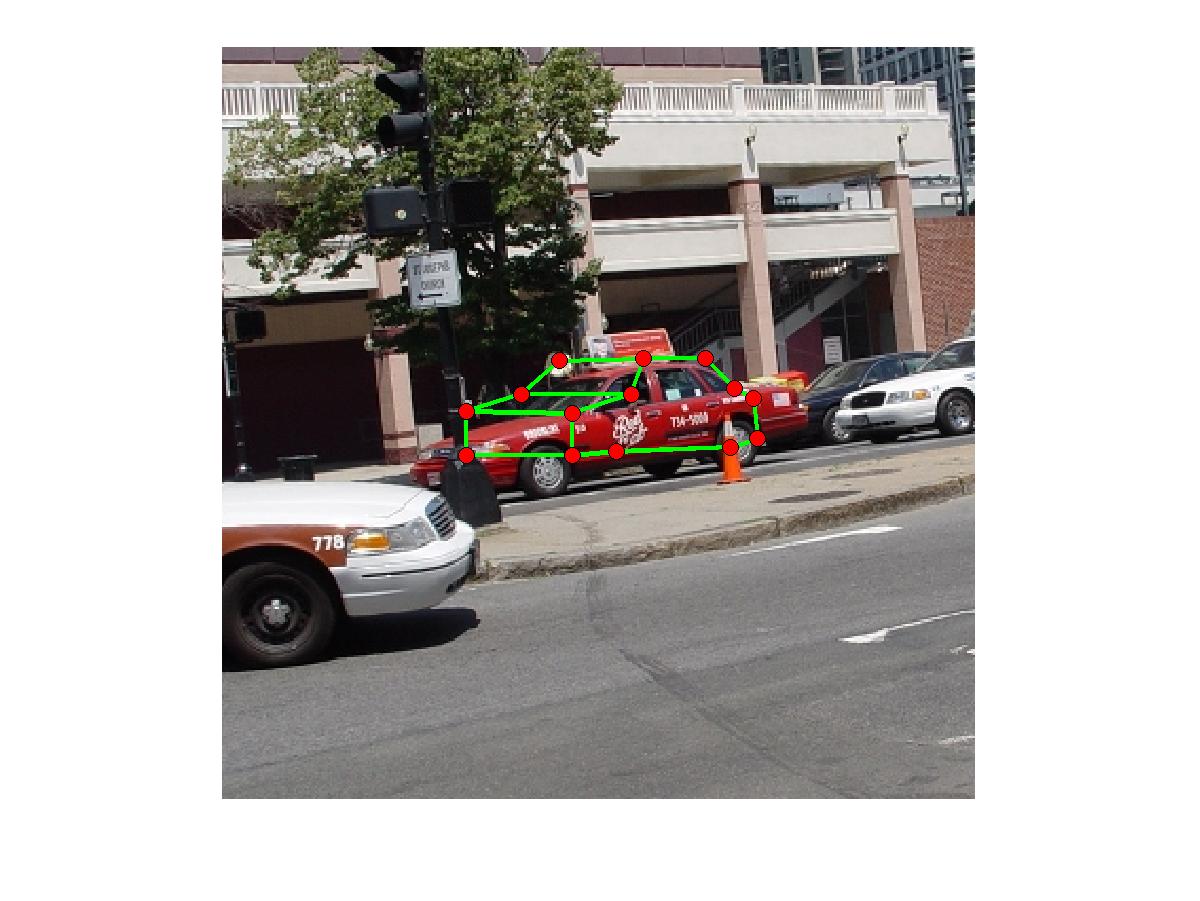}	
	\caption{Car Alignment Example 1) Top Row: MMVCF landmark detector 2) Bottom Row: VCF landmark detector}
	\label{fig:alignment_examples}
\end{figure}

Quantitatively the performance of the different landmark detectors is evaluated by computing the root mean square error (RMSE) of the detected landmarks with respect to manually labeled ground truth landmark locations. More specifically we report the landmark-wise average RMSE. In Fig.\ref{fig:alignment_error} we show the landmark-wise RMSE comparison between the different landmark detectors. We observe that MMVCF improves landmark localization slightly in comparison to VCF and significantly improvement over SVMs across all the landmarks. The poor performance of SVMs in this case is due to the limited availability of training samples (less than 400 samples per landmark). MMVCF lowers the RMSE (cumulative RMSE over all the landmarks in the image) for 526 images (i.e., lower RMSE on 52) in comparison to VCF. While the difference between the alignment using VCF and MMVCF is quite small in most images, in Fig.\ref{fig:alignment_examples} we show qualitative alignment results on some images where VCF fails spectacularly while MMVCF succeeds.
\subsection{PASCAL VOC}
   We evaluate the detection capability of the proposed classifier on a few classes (car, bus and bicycle) of the PASCAL VOC object detection dataset. We train whole object detectors using images from the PASCAL VOC 2012 challenge and evaluate the detection performance on the test set of PASCAL VOC 2007. The main idea behind MMVCF is to improve object localization performance by forcing sharper peaks in the correlation outputs. Therefore we use larger images for training the object detectors since not much can be gained by forcing sharper peaks on small templates. Further correlation filters by virtue of forcing sharp peaks in the correlation plane implicitly assume that the center of the training image is the center of the object, unless this information is explicitly provided. Since the training annotations of PASCAL VOC are weak from this perspective, for training, we only use images which are not labeled as difficult, truncated or occluded in the training set and use the validation set to cross-validate on the best parameters. We cluster the data in each into 3 mixtures using aspect ratio as in done in Deformable Parts Model (DPM) \cite{felzenszwalb2010object}. We perform several rounds of hard negative mining to train all the detectors. In addition we also learn DPM \footnote{\url{http://cs.brown.edu/~pff/latent-release4/}} root models from the same positive training images as a comparison. Table \ref{table:pascal} shows the average precision evaluation of our object detectors. We observe that MMVCF improves the average precision both over SVMs and over VCFs (by a very large margin). The margin maximizing constraints of both SVM and MMVCF provide better tolerance to outliers in the large amount of training samples available in comparison to VCF resulting in better object detection performance under the regime of large amounts of data with outliers. Therefore MMVCF outperforms both SVM and VCF under the regime of large scale data as well. Further the improvement object localization performance at the object as well as the parts level suggests that there may be room for improvement by replacing the SVM with MMVCF in the Deformable Parts Model.
   \begin{table}[!h]
      \centering
      \caption{PASCAL VOC 2007: Average Precision (in \%)}
      \label{table:pascal}
      \begin{tabular}{|c|c|c|c|c|}
      \hline
      {\footnotesize Object Class} & {\footnotesize DPM Root} & {\footnotesize VCF} & {\footnotesize SVM ($\gamma=1$)} & {\footnotesize MMVCF}\\
      \hline
       Car & {\footnotesize 44.9} & {\footnotesize 35.6} & {\footnotesize 43.9} & {\footnotesize 48.4} \\
       \hline
       Bus & {\footnotesize 40.2} & {\footnotesize 33.7} & {\footnotesize 40.5} & {\footnotesize 42.9} \\
       \hline
       Bicycle & {\footnotesize 40.9} & {\footnotesize 36.0} & {\footnotesize 39.4} & {\footnotesize 42.5} \\
      \hline
      \end{tabular}
   \end{table}
\subsection{MITStreetScene Cars}
\begin{figure*}[!ht]
   \centering
   \subfigure[View 1]{\includegraphics[scale=0.75]{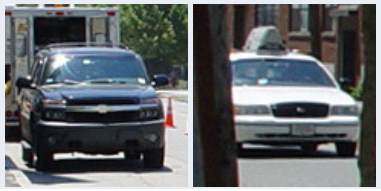}}
   \subfigure[View 2]{\includegraphics[scale=0.75]{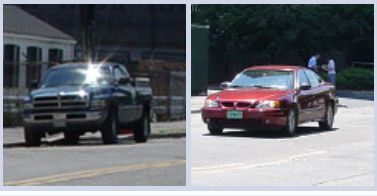}}
   \subfigure[View 3]{\includegraphics[scale=0.75]{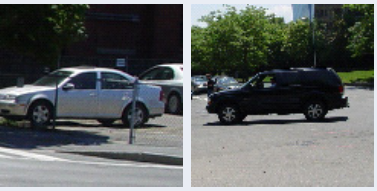}}
   \subfigure[View 4]{\includegraphics[scale=0.75]{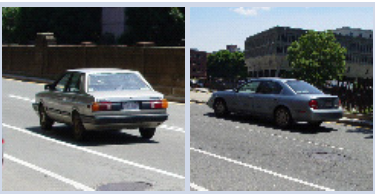}}
   \subfigure[View 5]{\includegraphics[scale=0.75]{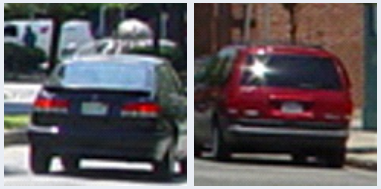}}
   \caption{Example training images from MIT StreetScene dataset for 5 different viewpoints}
   \label{fig:car-examples}
\end{figure*}
\begin{figure*}[!ht]
   \centering
   \subfigure{\includegraphics[scale=0.14]{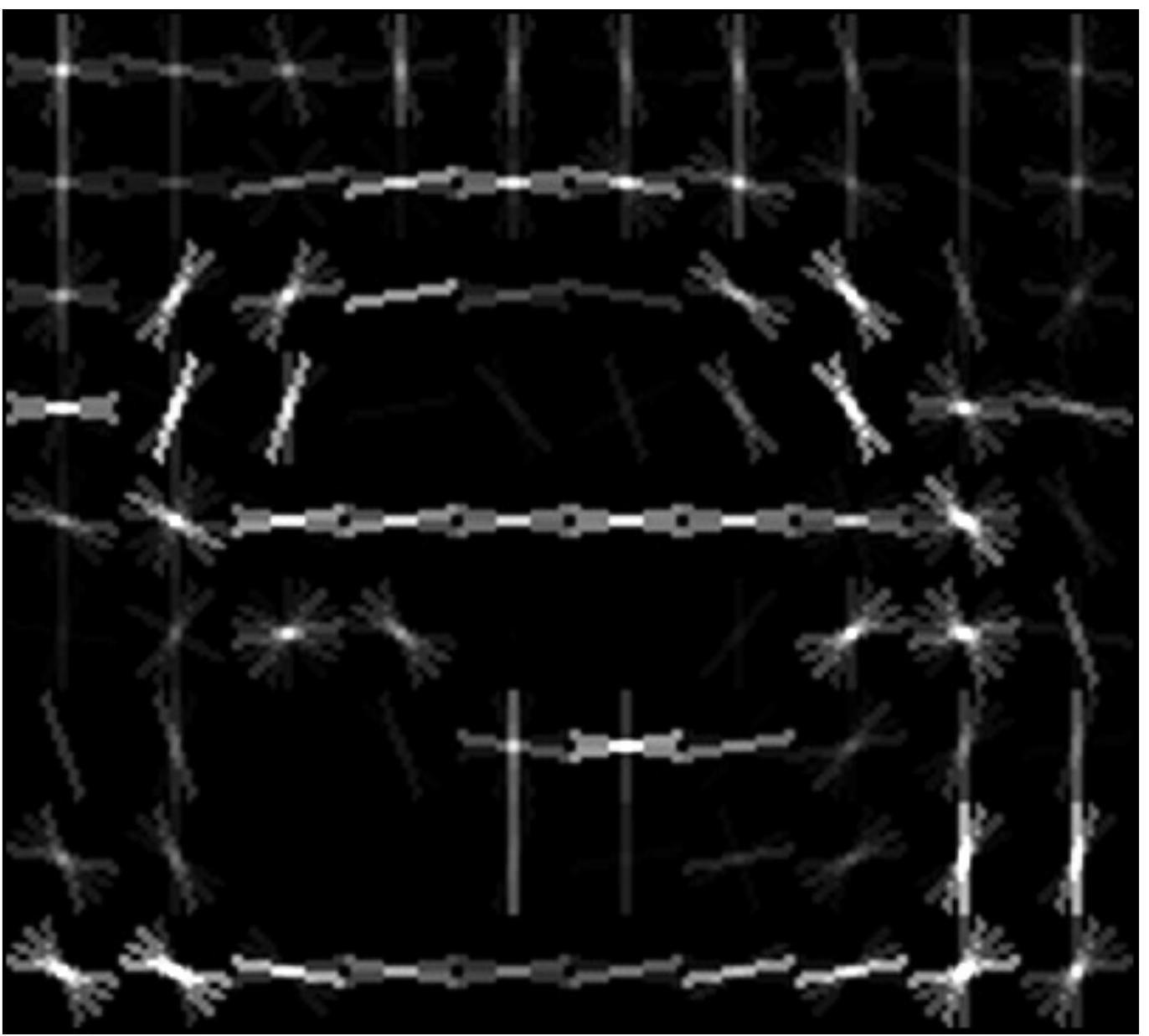}}
   \subfigure{\includegraphics[scale=0.20]{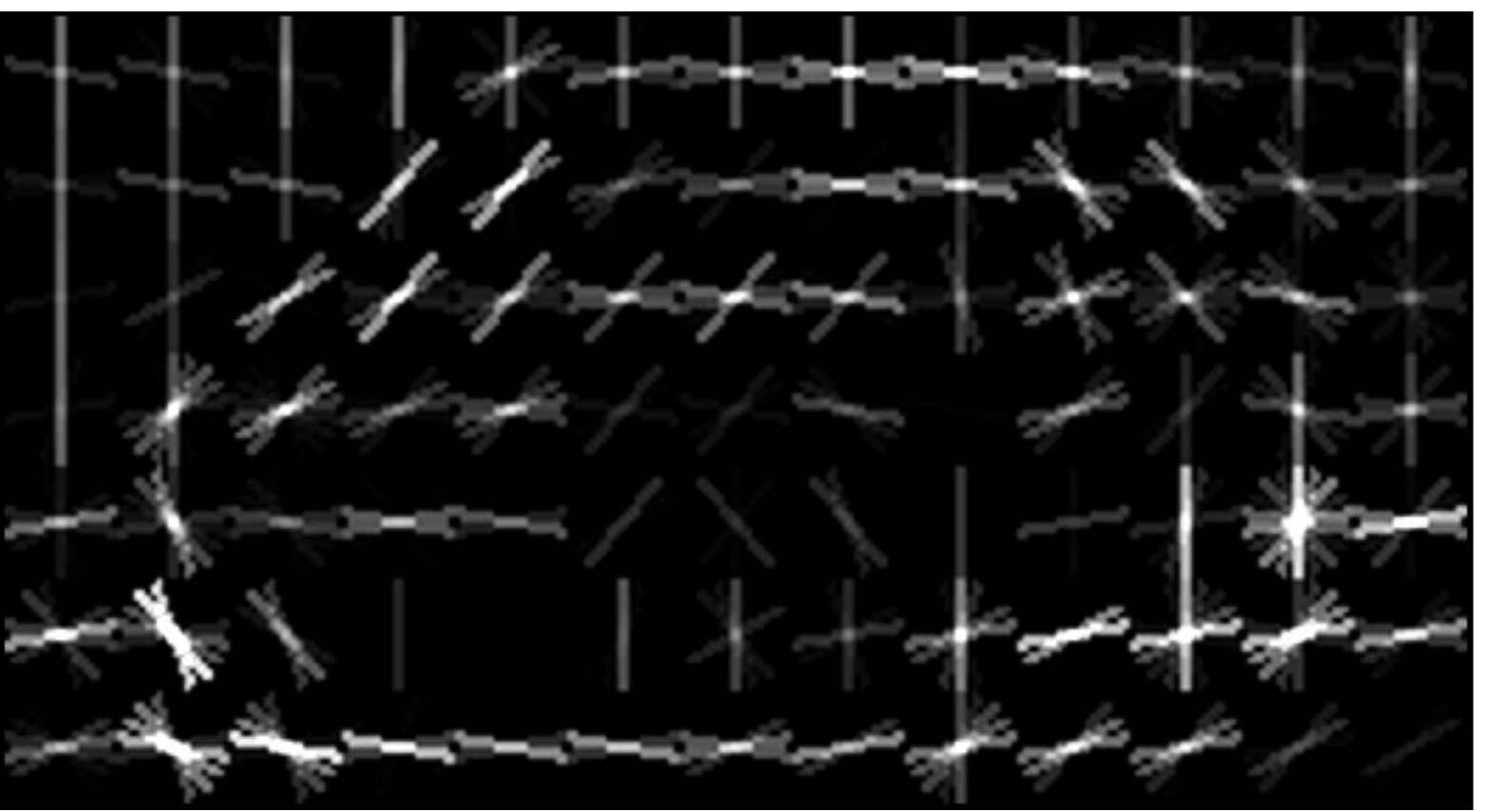}}
   \subfigure{\includegraphics[scale=0.27]{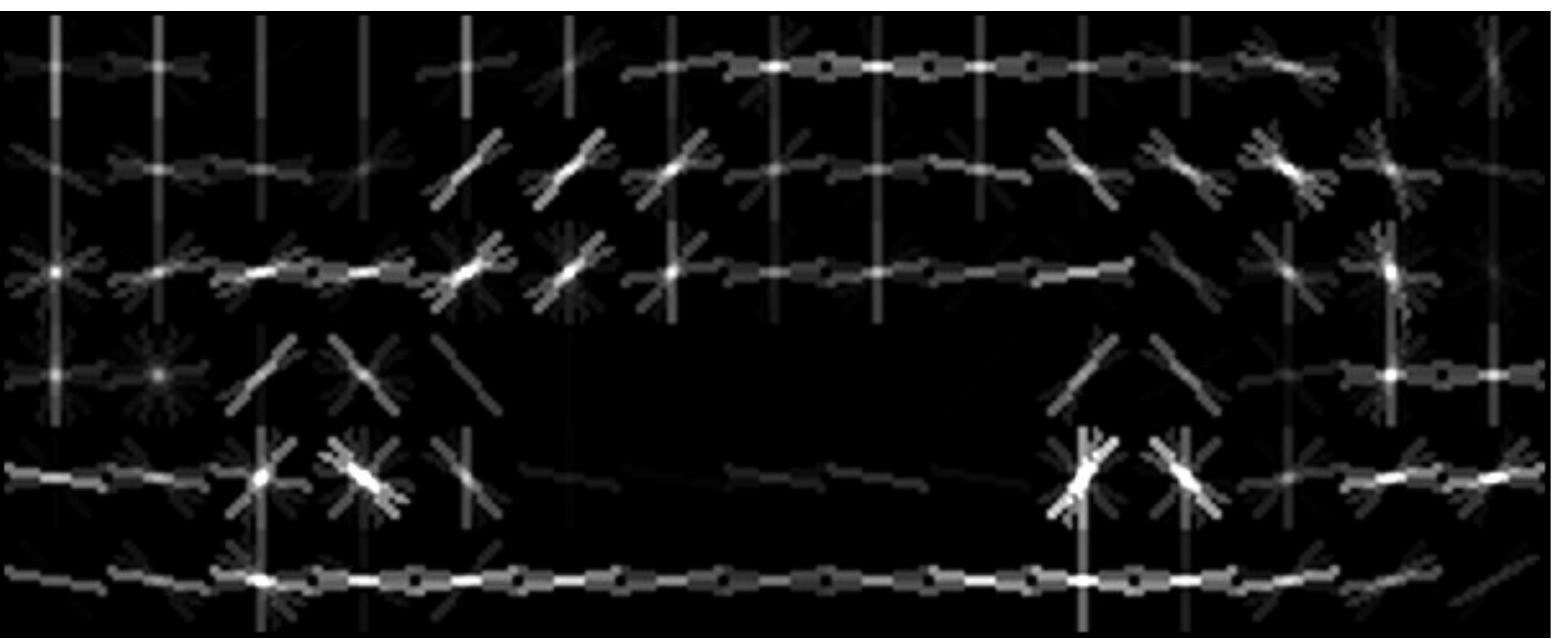}}
   \subfigure{\includegraphics[scale=0.20]{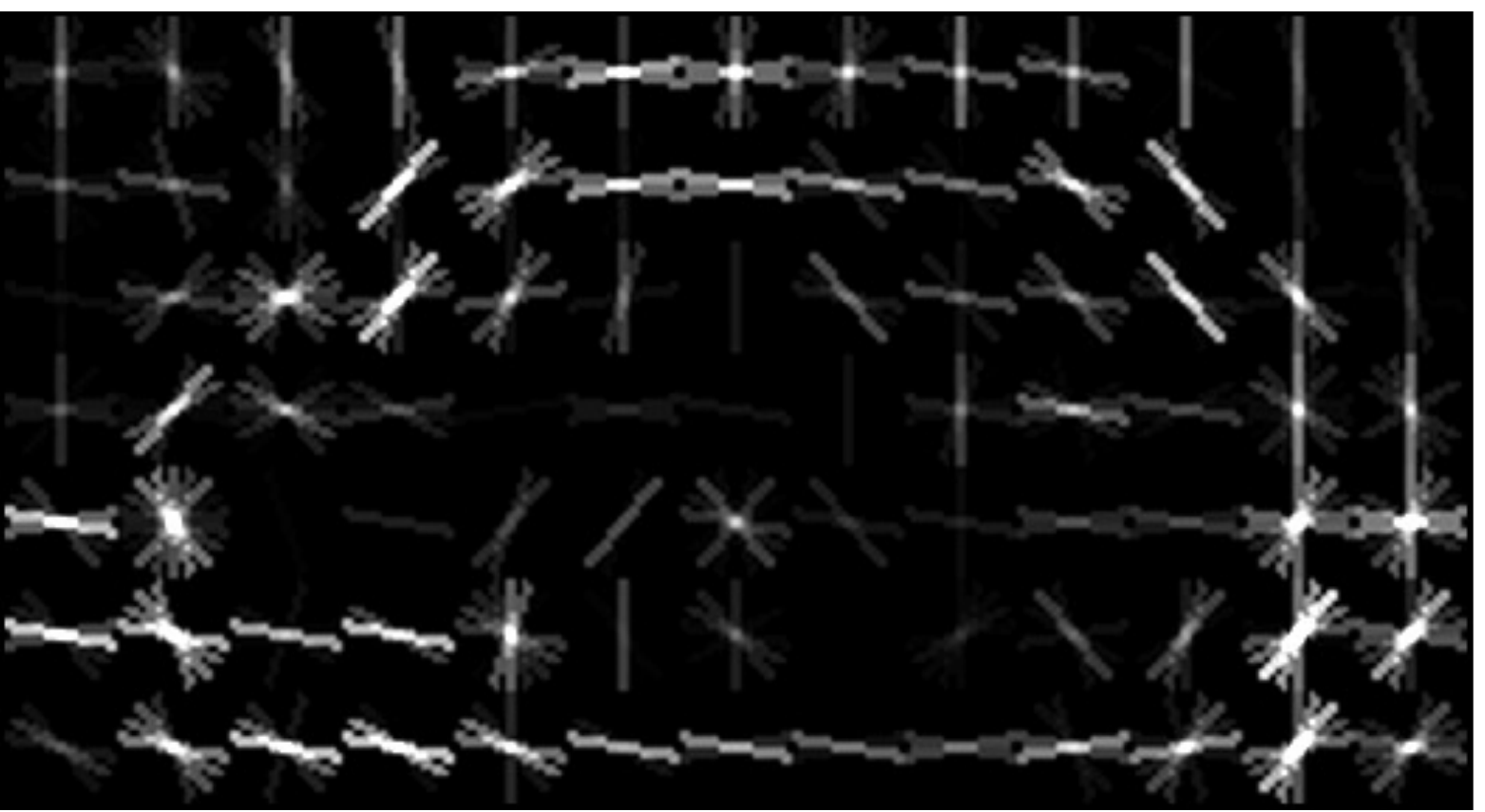}}
   \subfigure{\includegraphics[scale=0.14]{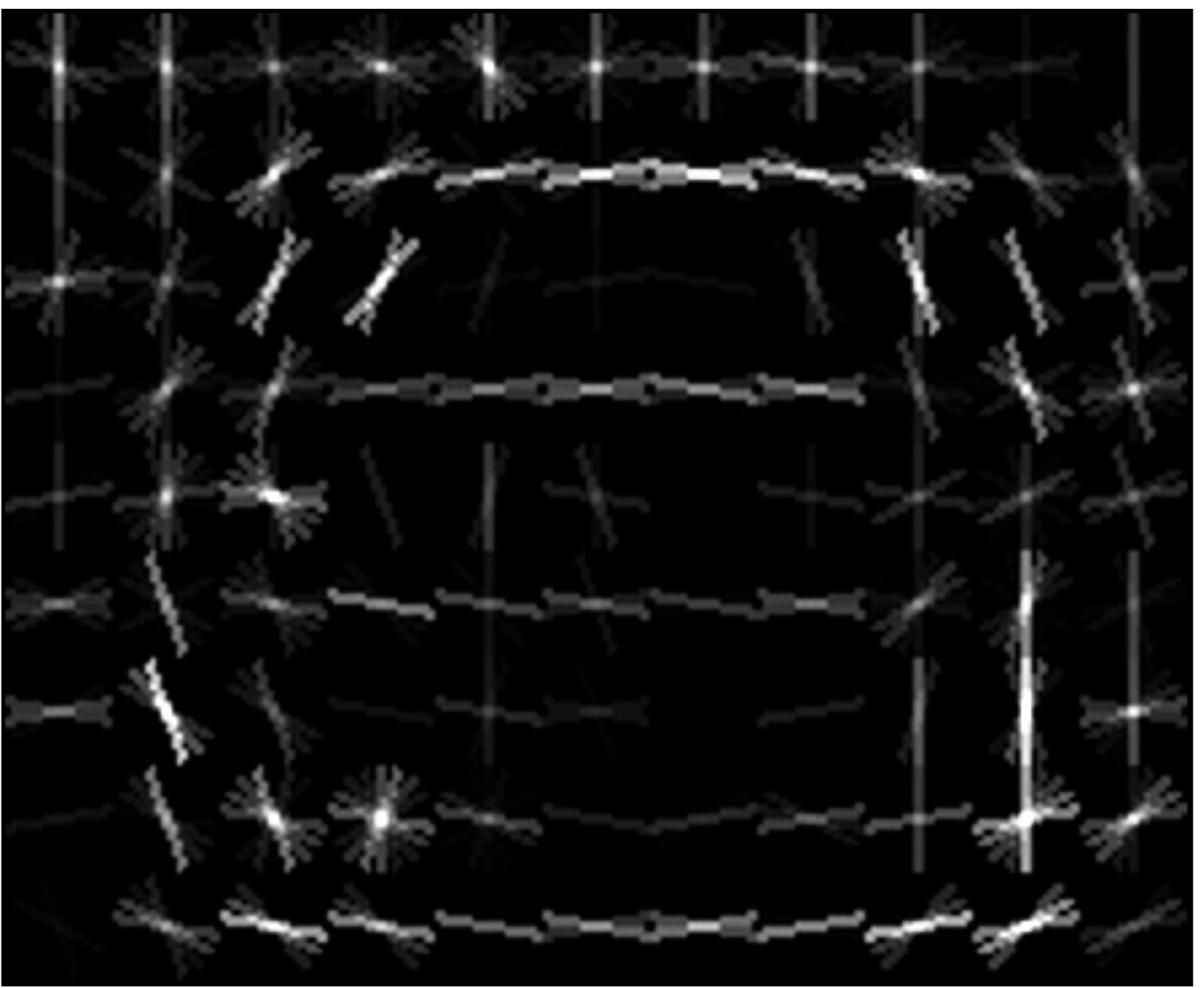}}\\
   \subfigure{\includegraphics[scale=0.14]{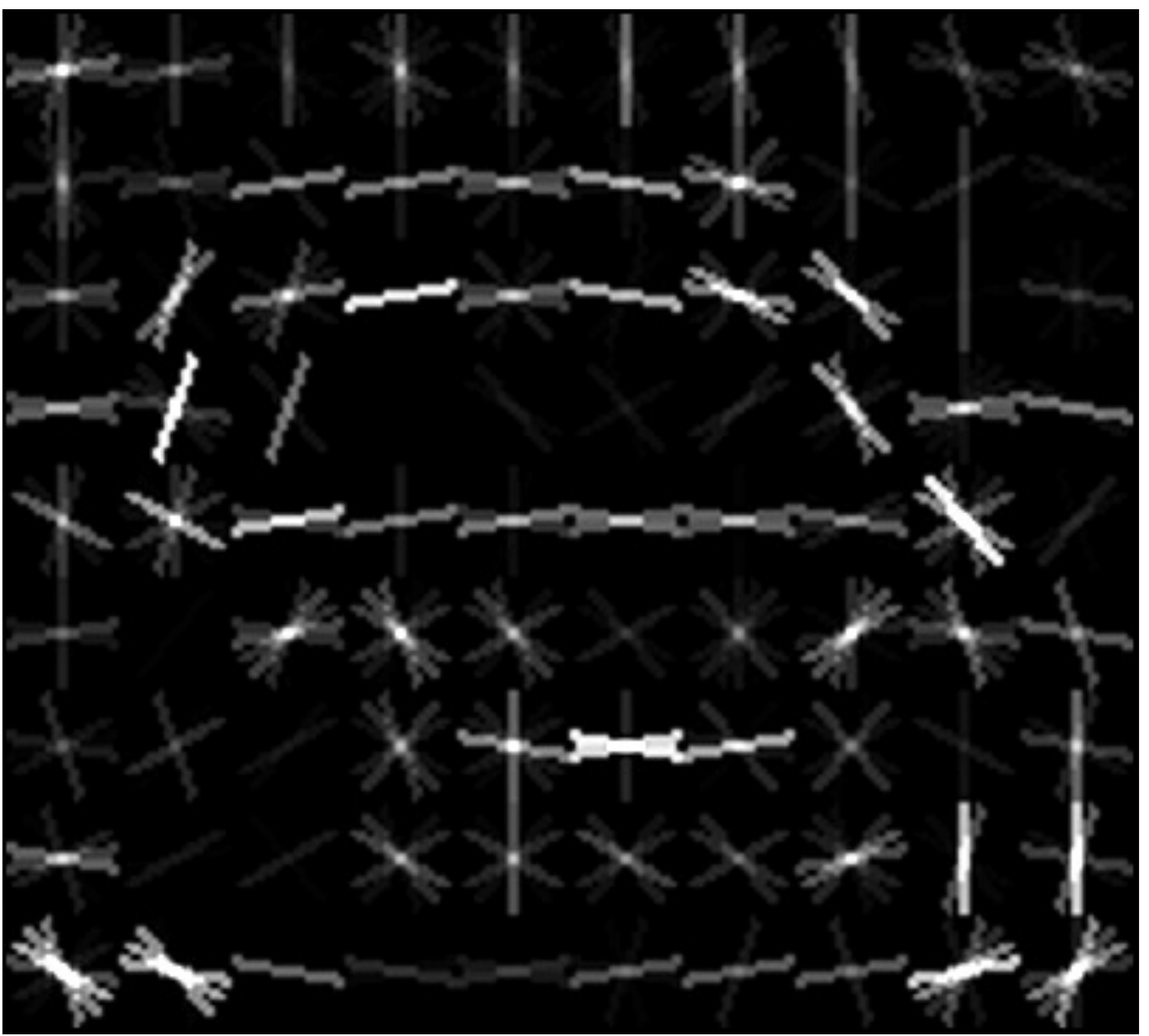}}
   \subfigure{\includegraphics[scale=0.20]{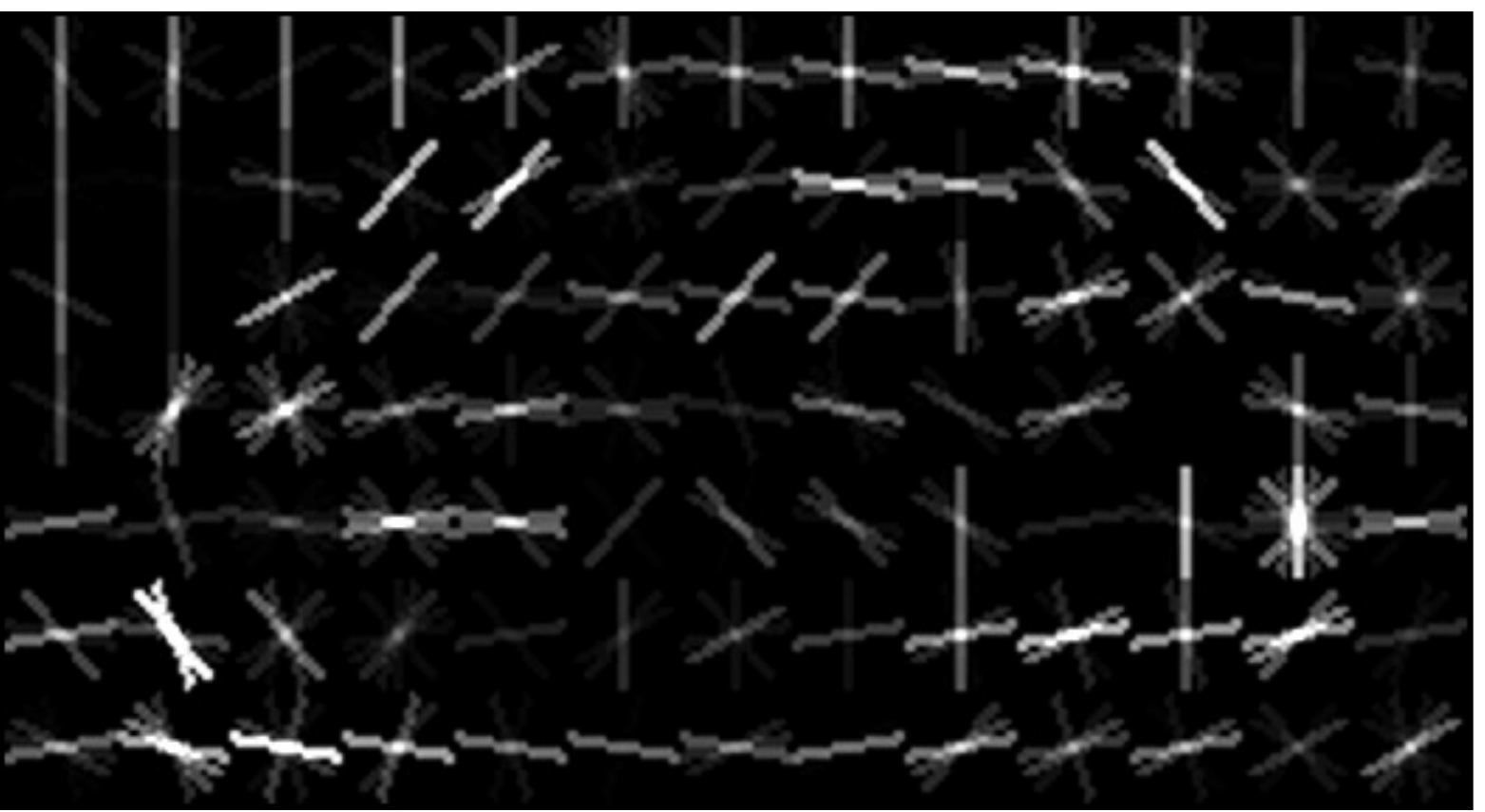}}
   \subfigure{\includegraphics[scale=0.27]{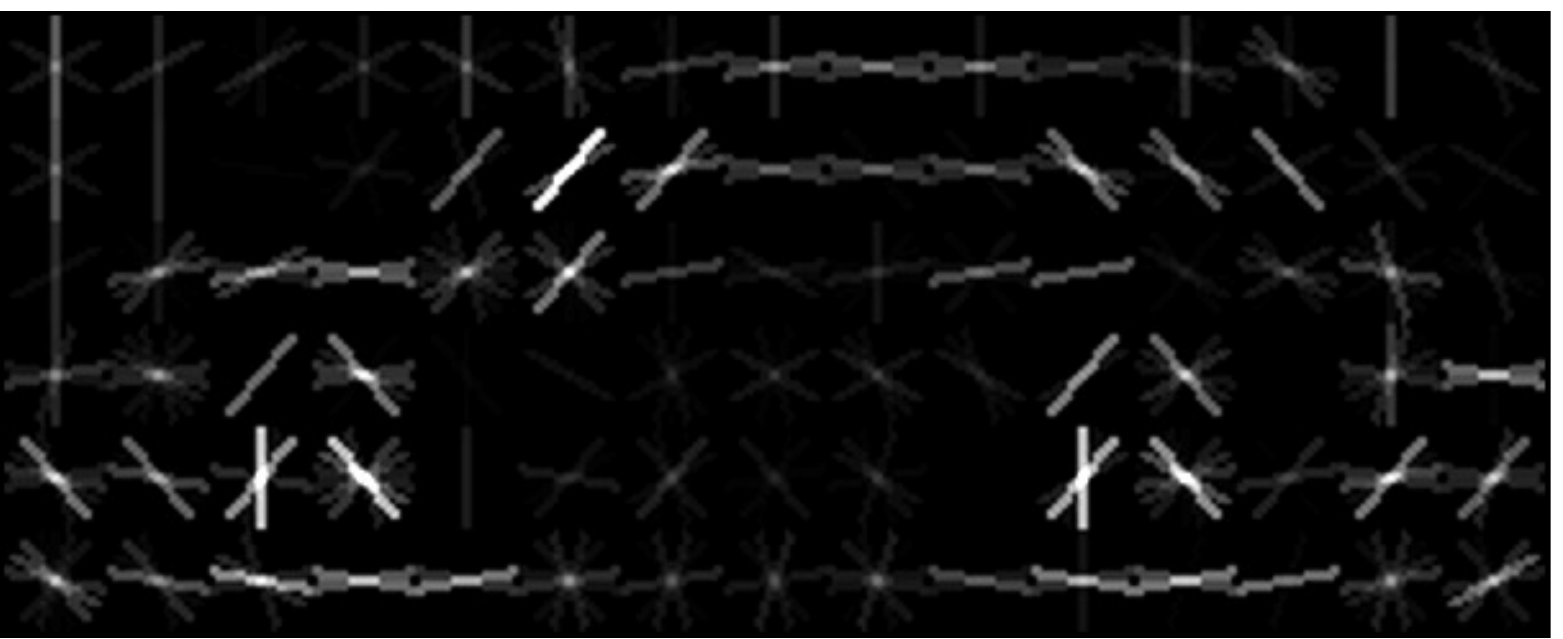}}
   \subfigure{\includegraphics[scale=0.20]{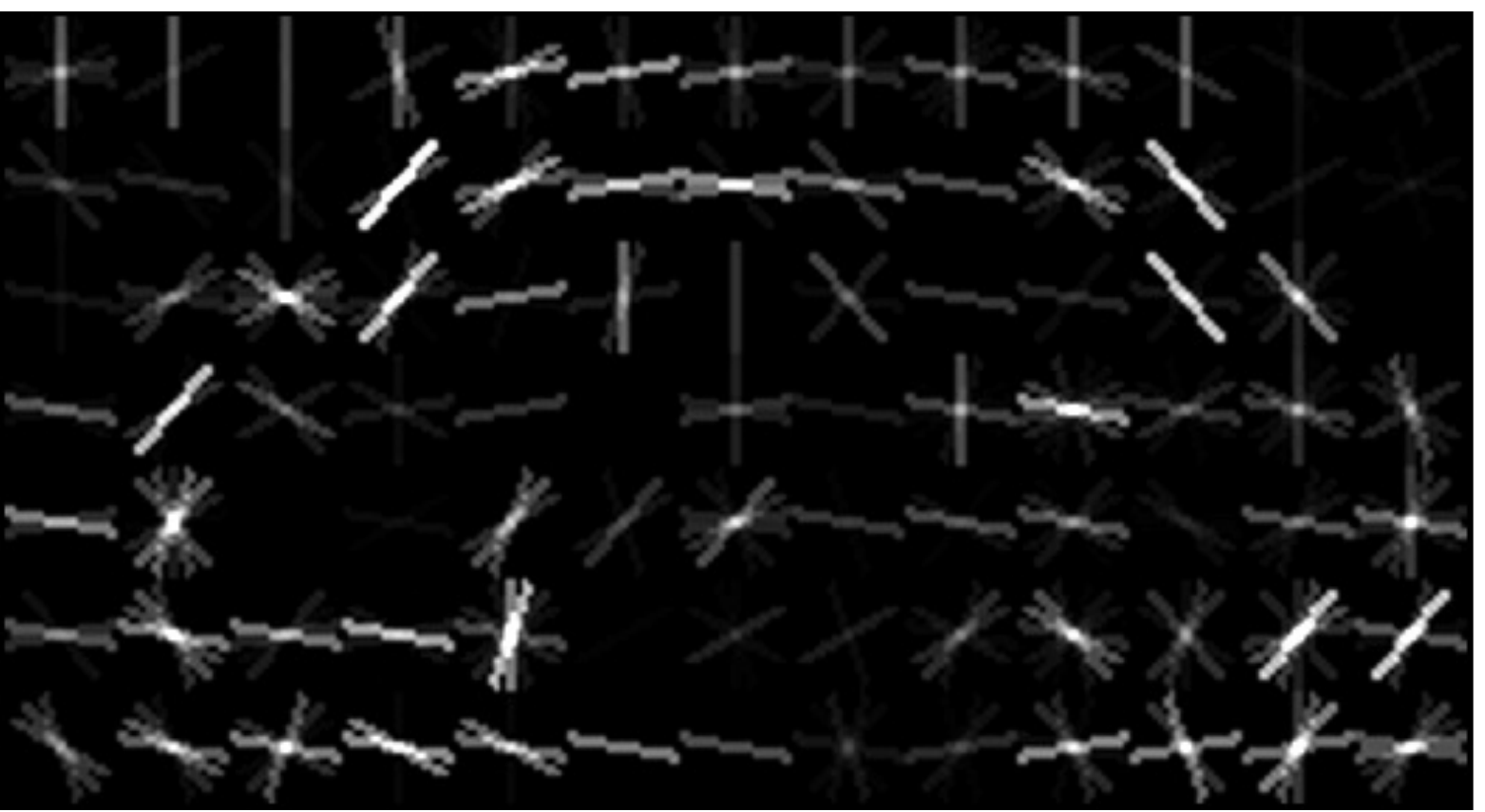}}
   \subfigure{\includegraphics[scale=0.14]{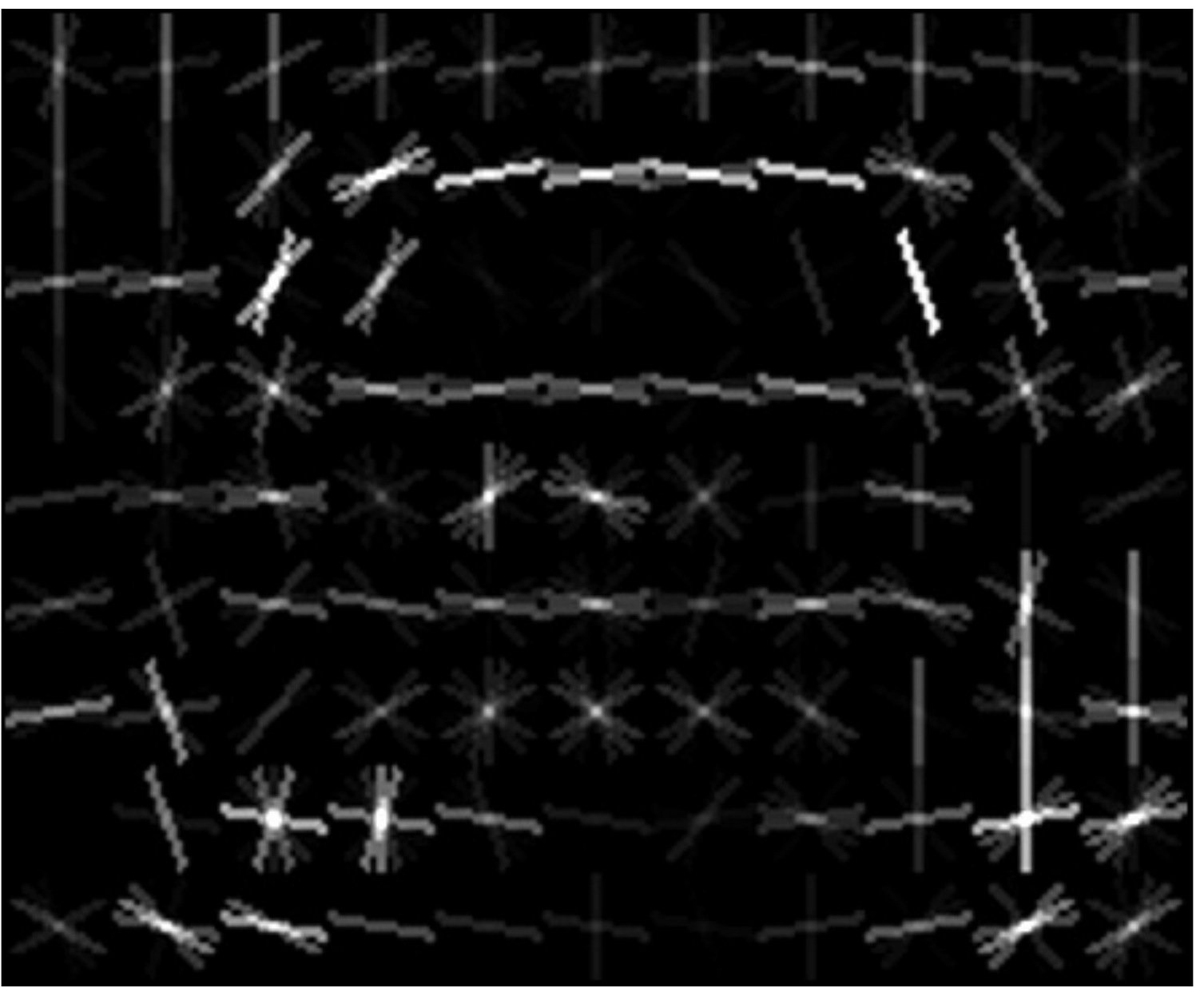}}
   \caption{Visualization of the learned root filters for five car poses. Top Row: HOG SVM. Bottom Row: HOG MMVCF.}
   \label{fig:filters}
\end{figure*}
Finally we consider the challenging scenario of training and testing the proposed object detector on different datasets. We design multi-view car detectors for detecting cars in unconstrained scenes using training samples from the MIT StreetScene dataset \cite{mit-street-dataset}. This dataset contains 3,547 street scene images which were originally created for the task of object recognition and scene understanding under an uncontrolled environment. For training we used 3,433 labeled cars which span a wide variety of types, sizes, background scenes, lighting conditions but excluding cars which are only partially visible. The images are manually segregated into 5 different poses as shown in Fig. \ref{fig:car-examples}. As a result we train 2 templates (original image and its horizontal mirror flip image) per view for a total of 10 templates. We first train a car detector using the Deformable Parts Model (DPM) \cite{felzenszwalb2010object} \footnote{\url{http://cs.brown.edu/~pff/latent-release4/}} (root filter only and root+parts). Using the exact training images used by the DPM Root filter (both positive and mined negative images) we train the MMVCF to enable a direct comparison between the SVM and MMVCF based classifier formulation when trained using the same exact training images. A detection is declared when the bounding boxes overlap by a factor of more than 0.5 and we report the Average Precision (AP). We test the car detector models trained above i.e., DPM Root, DPM Full (Root+Parts), SVM Root ($\gamma=1$ in the MMVCF formulation) and MMVCF Root on the following datasets. Parameters for our MMVCF formulation are estimated via cross-validation on a small subset of images from the LabelMe \cite{russell2008labelme} dataset.
\begin{enumerate}
   \item We first evaluate our approach on a dataset compiled by Hoiem et al. \cite{hoiem2008putting} which contains 422 random outdoor images from the LableMe dataset for a total of 923 cars. Those images cover a multitude of outdoor urban scenes and include a wide variety of object pose and size, making the dataset very challenging.
   \begin{table}[!h]
      \centering
      \caption{LabelMe (Hoeim et.al.): Average Precision}
      \label{table:derek}
      \begin{tabular}{|c|c|c|c|c|}
      \hline
      {\footnotesize A.P.} & {\footnotesize DPM Root} & {\footnotesize DPM Full} & {\footnotesize SVM Root} & {\footnotesize MMVCF Root}\\
      \hline
      {\footnotesize (in \%)} & {\footnotesize 37.9} & {\footnotesize 39.4} & {\footnotesize 37.4} & {\footnotesize 41.6} \\
      \hline
      \end{tabular}
   \end{table}
   \item We also evaluate the trained car detectors on the car category of the PASCAL Visual Object Classes (VOC) Challenge 2007 \cite{pascal-voc-2007} dataset. We report the AP for the car category in Table \ref{table:pascal} following the PASCAL VOC 2007 evaluation protocol. Note that these results were achieved \emph{without} using any positive training examples from PASCAL VOC dataset. The disparity in car detection performance between training the detectors on the MITStreetScene dataset and the PASCAL VOC 2012 dataset is likely due to the problem of dataset bias \cite{torralba2011unbiased}.
   \begin{table}[!h]
      \centering
      \caption{PASCAL VOC 2007: Average Precision}
      \label{table:pascal}
      \begin{tabular}{|c|c|c|c|c|}
      \hline
      {\footnotesize A.P.} & {\footnotesize DPM Root} & {\footnotesize DPM Full} & {\footnotesize SVM Root} & {\footnotesize MMVCF Root} \\
      \hline
      {\footnotesize (in \%)} & {\footnotesize 35.1} & {\footnotesize 40.5} & {\footnotesize 35.4} & {\footnotesize 39.0} \\
      \hline
      \end{tabular}
   \end{table}
\end{enumerate}
\section{Discussion}
The localization loss criterion in the template learning formulation induces a linear similarity function, $\mathbf{\hat{x}^{\dagger}_{i}}\mathbf{\hat{S}^{-1}\hat{x}_{j}}$, between samples $\mathbf{x}_i$ and $\mathbf{x}_j$ where $\mathbf{\hat{S}}$ is the cross-power spectrum of the training samples capturing the second order statistics of the data. This weighted dot product, with the weights being the inverse cross-power spectrum of the samples, accounts for the \emph{correlations between the samples and their respective geometrically shifted (spatial translation) versions} captured by the diagonal elements of $\mathbf{\hat{D}}$ as well as the \emph{pairwise correlations across the different feature channels} captured by the non-diagonal elements of $\mathbf{\hat{D}}$. Note that while SVMs have long been used with vector features like Gabor filter banks and HOG, the linear SVM kernel $\mathbf{x}_i^T\mathbf{x}_j$ (equivalently $\mathbf{\hat{x}}_i^{\dagger}\mathbf{\hat{x}}_j$ in the frequency domain) does not explicitly account for correlations across the different feature channels. By accounting for the redundancies across the feature channels the MMVCF formulation effectively has more degrees of freedom enabling it to model more complex functions in comparison to the linear SVM kernel. Although the presence of multiple feature channels helps improve the generalization capability of the correlation filters noisy or corrupt data often found in real world large scale vision datasets can hurt the performance of the filter like VCF which are not explicitly designed to handle outliers. The margin maximizing formulation of MMVCF, which is known to promote generalization, helps to mitigate this limitation of VCFs leading to improved generalization capability over VCF and improved localization capability over SVMs.

\section{Conclusion}
Conventional CFs are not designed to be used with vector feature representations. Recently correlation filter designs for vector-valued or multi-channel features have been proposed with attractive computational and memory efficiencies but seem to suffer from limited robustness to outliers in noisy data. In this paper we introduced the Maximum Margin Vector Correlation Filter (MMVCF) which is a correlation filter design for multi-channel features which combines the attractive localization properties of traditional correlation filter designs and the generalization and robustness capabilities of margin maximizing classifiers like SVMs. We evaluated this classifier on multiple datasets for the tasks of object detection and object alignment and demonstrated that MMVCF outperforms both SVMs and other correlation filter designs in the regimes of both small scale as well as large scale training samples.

{\small
\bibliographystyle{ieee}
\bibliography{mmvcf}
}

\end{document}